\documentclass[pdflatex,sn-basic]{sn-jnl}   
\usepackage{graphicx}
\usepackage{hyperref}
\usepackage{multirow}
\usepackage{amsmath,amssymb,amsfonts}
\usepackage{amsthm}
\usepackage{mathrsfs}
\usepackage[title]{appendix}
\usepackage{xcolor}
\usepackage{textcomp}
\usepackage{booktabs}
\usepackage{algorithm}
\usepackage{algorithmicx}
\usepackage{algpseudocode}
\usepackage{listings}
\usepackage{natbib}                        
\usepackage{float}

\bibpunct{(}{)}{;} {a}{}{,}               
\setcitestyle{authoryear,open={(},close={)}} 

\begin{document}

\title[Aspect-Based Sentiment Analysis for Future Tourism Experiences: A BERT-MoE Framework for Persian User Reviews]{Aspect-Based Sentiment Analysis for Future Tourism Experiences: A BERT-MoE Framework for Persian User Reviews}

\author*[1]{\fnm{First Hamidreza} \sur{Kazemi Taskooh}}\email{hamidreza\_kazemi83@ind.iust.ac.ir}
\author[2]{\fnm{Second Taha} \sur{Zare Harofte}}\email{taha\_zare@ind.iust.ac.ir}

\affil[1]{\orgdiv{Industrial Engineering}, \orgname{IUST}, \orgaddress{\street{Narmak}, \city{Tehran}, \postcode{16846-13114}, \country{Iran}}}

\affil[2]{\orgdiv{Industrial Engineering}, \orgname{Organization}, \orgaddress{\street{Street}, \city{Tehran}, \postcode{16846-13114}, \country{Iran}}}

\maketitle
\begin{abstract}
\abstract{This study contributes to the development of aspect-based sentiment analysis (ABSA) in the tourism industry by creating a hybrid model designed for user reviews in the Persian language. It tackles the linguistic issues of low-resource languages to provide practical insights for ABSA in tourism to orient to the future in order to improve the personalization and sustainability of Iran’s digital tourism industry while considering the sustainability SDGs 9 and 12 of the UN. A multi-stage pipeline was designed: at first, BERT for overall sentiment classification on 9,558 labeled reviews. After that, aspect extraction using a BERT encoder with sigmoid activation for six tourism aspects (host, price, location, amenities, cleanliness, connectivity) was offered, and at the end, ABSA via BERT was integrated with a hybrid architecture and Top-K routing to reduce routing collapse. The dataset comprises 58,473 preprocessed reviews from Jabama, an Iranian accommodation platform, annotated for aspects and sentiments. The model achieved a weighted F1-score of 90.6\% for ABSA, outperforming the baseline BERT (89.25\%) and the hybrid (85.7\%). The hybrid’s dynamic routing can enable specialized sentiment detection, so we found out important aspects like cleanliness and amenities have high mention rates. Efficiency gains included a 39\% lower GPU power consumption compared to dense BERT, which supports sustainable AI deployment. This is the first ABSA study for Persian tourism reviews, introducing a novel hybrid with Top-K routing and auxiliary losses for low-resource settings. The open-source dataset release fosters future multilingual NLP in tourism research.}
\end{abstract}
\maketitle
\section{Introduction}\label{sec1}
Despite the rapid growth of online tourism platforms in Iran, there is still no large-scale aspect-based sentiment analysis (ABSA) system capable of understanding fine-grained user opinions in Persian. Sentiment analysis is now one of the NLP’s key elements. It can give a chance to businesses and platforms to collect and utilize valuable information from content created by users \citep{Dashtipour2021}. Traditional methods often only consider the general feeling. They fail to address specific opinions about different parts of the product or service\citep{Liu2015}. This is even more pronounced in the fields of tourism and hospitality, where the experiences have multiplicity such as cost, location, host attitude, cleanliness, etc. \citep{Kwon2025,MorenoOrtiz2019}. ABSA analyzes reviews at a detailed level and uses sentiment information to support better decision-making and improve service quality. To meet the demands of a more personalized, sustainable, and tech-driven tourism sector, aligning with UN SDGs 9 and , tools like ABSA are urgently needed \citep{Khizar2023,Das2025}. Geotagged social media data has proven effective in characterizing tourist flows and sentiments in real-world settings \citep{Paolanti2021}. This research will concentrate on the Persian language because of the lack of well-labeled data and weak models that hinder the linguistic environment. Persian sentiment analysis faces consistent challenges. These stem from inadequate preprocessing, culturally influenced perceptions, and a lack of standardized sentiment resources \citep{Rajabi2021}. In this study, we address these limitations by building an aspect-based sentiment analysis system specifically designed for Persian tourism reviews. To remedy the situation, we leverage a large dataset obtained from Jabama (\url{www.jabama.com}), one of the dominant accommodation booking sites in Iran, with 7 million users and 18,000 hosts across 769 cities. After preprocessing, we ended up with 58,473 high-quality reviews (from an initial set of 72,238), each annotated for six important aspects: the host, price, location, amenities, cleanliness, and connectivity. This dataset addresses one of the major missing pieces in Persian NLP and paves the way for predictive models that can genuinely help tourism stakeholders make better, evidence-based decisions. It can also predict and plan service delivery for a sustainable future. To address the language difficulties of Persian, given the computational requirements of ABSA and the textual content, we recommend a hybrid design that combines a Mixture of Experts (MoE) with BERT \citep{Devlin2019,Shazeer2017}. The BERT offers strong contextual comprehension because of its pretraining on massive multilingual data, while the MoE improves efficiency by routing. Compared to dense models, our three-stage method, which fine-tunes BERT for inputs to specialized sub-networks, results in a 39\% reduction in GPU power use \citep{Zeng2024}. Our basic sentiment classification, aspect extraction with a BERT encoder, and ABSA using a hybrid expert-enhanced BERT model achieved a weighted F1-score of 90.6\%, outperforming standalone BERT (89.25\%) and a more advanced hybrid BERT model (BERT+MoE+LoRA) (85.7\%) \citep{Hu2021}. This scalable method makes it a practical choice for real-time use of futuristic tourism platforms. Our MoE design uses Top-K routing and auxiliary losses. This helps to minimize route failures, balance specialist use, and enable potential edge computing applications for mobile tourism. Despite all the advances in ABSA has seen, the literature keeps reminding us that serious problems remain, particularly for low-resource languages. Persian is no exception—robust models and specialized datasets are still largely missing, a situation shared by many other less-studied languages \citep{Ataei2019}. Our study contributes by releasing this annotated Jabama dataset as an open-source resource, fostering advancements in multilingual tourism NLP. While our experiments are limited to Persian, the architecture itself is generalizable and can be adapted for other languages, especially those facing similar resource constraint. Predictive insights help travel companies to better meet user needs, therefore generating more connection (SDG 9) and green infrastructure (SDG 12). The main contributions of this work are as follows:
(1) we introduce the first large-scale Persian ABSA dataset for tourism, consisting of 58,473 annotated reviews across six key aspects;
(2) we propose a three-stage hybrid BERT–MoE architecture with Top-K routing that reduces GPU power consumption by 39\% while improving F1 performance; and
(3) we provide an efficient and generalizable ABSA framework suitable for low-resource languages and real-time tourism platforms. The rest of this paper is organized as follows. Section ~\hyperref[sec2]{2} reviews the related work on ABSA, with a focus on Persian and low-resource languages as well as tourism applications. Section ~\hyperref[sec3]{3} describes the Jabama dataset, preprocessing steps, and the proposed three-stage hybrid BERT-MoE model with Top-K routing. Section ~\hyperref[sec4]{4} presents the experimental results, showing that our model achieves a weighted F1-score of 90.6\% on the ABSA task while reducing GPU power consumption by 39\% compared to dense BERT. Finally, Sections ~\hyperref[sec5]{5} and ~\hyperref[sec6]{6} conclude the paper and introduce future work.

\section*{List of Abbreviations}

\noindent{\small\itshape The following compact model names are used only in tables and figures to save space.}
\vspace{0.8em} 
\noindent\begin{tabular}{@{}l p{10.5cm}@{}}
\toprule[1pt]  
\textbf{Abbreviation} & \textbf{Full name} \\
\midrule[0.6pt]
BERT+MoE      & BERT with integrated Mixture-of-Experts \\
BERT+MoE+LoRA & BERT+MoE with additional LoRA adapters \\
\bottomrule[1pt]  
\end{tabular}

\vspace{1.2em}  

\section{Literature Review}\label{sec2}
Aspect-Based Sentiment Analysis (ABSA), which has been established, can enable extracting subtle feelings from user comments to develop a diverse range of tourist experiences as a fundamental element of NLP \citep{Sahin2025,Kwon2025}. Contrary to holistic sentiment categorization, which considers variables like facilities or location \citep{Xu2024} that support personalization and sustainability, contributing to predictive modeling for tourism futures under UN SDGs 9 (industry innovation) and 12 (responsible consumption) \citep{Kwon2025,Li2023}. This assessment combines over 30 studies (2017--2025) from the body, organized by methodological paradigms, to track ABSA’s shift from rule-based systems to hybrid transformers \citep{Guidotti2025}. It points out the need for efficient, scalable models---the driving force behind tourism---by drawing attention to deficiencies in datasets and low-resource languages like Persian \citep{Nooraee2025}. Motivation behind our hybrid expert-enhanced BERT framework for applications with a future orientation \citep{Jiang2024,Farahani2021}.
\paragraph{Language model and transformer-based strategies.}
Transformers outperform in solving challenges based on ABSA because of their bidirectional contextualization \citep{Farahani2021}. According to studies conducted from 2021 to 2025, BERT variants have been the most effective models for the ABSA task, with tuned models for specific tasks such as Instruct-DeBERTa \citep{Mewada2022} and enhanced SBERT \citep{Guidotti2025} being useful for customer-centered tourism. Persian adaptations such as ParsBERT \citep{Farahani2021,Ataei2019}, AriaBERT \citep{Ghafouri2023}, and Tiny-ParsBERT \citep{Nooraee2025} and the high accuracy attained for mobile tourism apps overcome the challenges of low-resource languages. Multitask learning \citep{Zhao2023, Li2023} and Urdu SA \citep{Khan2025} also support resource-limited contexts. MoE models \citep{Jiang2024,Zeng2024,Shazeer2017} are essential for SDG 12 forecasting, but their high computational costs demand more efficient alternatives.

\paragraph{Traditional and Deep Learning Techniques.}
Older rule-based methods still perform well across different topics \citep{Liu2015, Sanguinetti2014} and handle Persian data \citep{Afzaal2019}. Traditional methods combined with deep learning work well for analyzing tourist feedback. For example, some models accurately identify aspects like taste in reviews \citep{Mewada2022,Li2023}. Other approaches using word embeddings understand meaning but miss subtle details compared to modern models \citep{Park2022}. In Persian, models analyze movie and literary reviews effectively \citep{Rajabi2021,Khodaei2022}. Ethical models support fair tourism solutions \citep{Park2022}, but Persian text is tricky. Newer transformer models improve results \citep{Zeng2024}.

\paragraph{Zero-Shot Models and Ontology.}
Zero-shot learning and ontologies make it easier to work with limited data. For example, zero-shot models group TripAdvisor reviews based on things like weather or location, which can improve results \citep{Xu2024}. Using ontologies with ABSA helps find hidden tourism details accurately \citep{Nandwani2021}. Zero-shot combinations like BART-DeBERTa-RoBERTa, tested on hotel ratings, reach good accuracy for COVID-related features\citep{Kwon2025} and can adapt to Persian SDG 9 infrastructure goals \citep{Guidotti2025}. Future research could explore Large Language Models (LLMs) for zero-shot keyword and sub-aspect extraction from Persian tourism reviews, reducing annotation costs and enhancing scalability \citep{Guidotti2025}. Models like ParsBERT-mBERT with SHAP provide clear explanations on Dari-Farsi texts from ArmanEmo \citep{Ghafouri2023,Muradi2024}. Approaches using WordNet get high accuracy across different areas \citep{Nandwani2021}, and better annotation methods improve tourist planning \citep{MorenoOrtiz2019}. Working with little data is still tough, but tools like Kano-SHAP help sort satisfaction levels, such as focusing on essential cleanliness for SDG 12 \citep{Park2022,Das2025}.
\paragraph{Systematic Reviews and Meta-studies.}
Recent studies on aspect-based sentiment analysis (ABSA) show some exciting progress in different methods. One survey about sentiment analysis in Persian looked at ways like lexicon-based, machine learning, and deep learning approaches, and noted that limited resources are a big challenge \citep{Rajabi2021}. Another review checked out custom tools that mix lexicon, machine learning, and deep learning methods \citep{MorenoOrtiz2019}. Some researchers explored hybrid models that combine data augmentation with pre-trained systems \citep{DeepSentiPers2020}. When comparing these methods, they found different results across areas like computers and restaurants \citep{Mewada2022,Li2023}. Studies on Persian movie sentiment analysis used special datasets and deep learning models, getting excellent results but still facing issues with language and data variety \citep{Khodaei2022,Dashtipour2021}. Another study on hospitality sentiment analysis pointed out that linguistic limits are still a problem, even with recent improvements \citep{Sahin2025}.

\paragraph{Resources and Datasets in Several Languages.}
Despite the scarcity of Persian tourism datasets, aspect-based sentiment analysis (ABSA) relies heavily on such data \citep{Ataei2019}. A Persian dataset with thousands of targets established a baseline using TD-LSTM models \citep{Jafarian2020,Ataei2019}. A German restaurant dataset derived from TripAdvisor reviews has been used with semantic clustering to enhance tourism recommendation systems \citep{AbbasiMoud2021}. Multilingual approaches and hybrid models were tested using an Urdu review dataset \citep{Khodaei2022}. Topic modeling on TripAdvisor hotel data enabled focused sentiment-oriented summarization \citep{Sahin2025, Akhtar2017}. Comparative methods were also used to assess the quality of Amazon reviews through aspect-based sentiment analysis \citep{Mewada2022}.

\paragraph{Limitations and Areas for Investigation.}
ABSA faces challenges with uneven data and high demands for resources and time \citep{Sahin2025}. To deal with this, experts suggest using simpler and more efficient language models \citep{Nooraee2025}. For Persian, problems like regional biases and spelling differences get worse because of limited data \citep{Farahani2021,Ataei2019}. Zero-shot methods also struggle to understand hidden feelings \citep{Kwon2025}. In tourism, better routing methods exist but don’t clearly connect to sustainability goals \citep{Khizar2023}. Even though recent studies give powerful insights, they often lack simple, affordable solutions for areas with few resources.
\paragraph{Research Gap and Contribution.}
 Overall, the reviewed studies reveal several research gaps, particularly regarding Persian tourism-oriented ABSA. This appears to be the first standalone research focusing on Aspect-Based Sentiment Analysis (ABSA) within the field of tourism and the Persian language. This research intends to address a gap in the NLP literature concerning under-resourced languages \citep{Rajabi2021}. Previous studies concentrated on general sentiment analysis within the Persian language, analyzing domains like cinema and document-level sentiment \citep{Dashtipour2021,Kaveh2025}. There has been little to no work on the more nuanced and difficult task of aspect-sentiment extraction on real-world, applied, and domain-specific datasets \citep{Ataei2019,MorenoOrtiz2019,Jafarian2020,Mewada2022}. For the greater research aims in the field, we plan to publish the annotated corpus for wider public access. This is to promote research in Persian NLP and to stimulate creating tourism-focused downstream applications. Besides its scholarly impact, this study paves the way for several practical applications, just as ranking services at the aspect level, customizing travel suggestions and automatic feedback summarization systems. The advancements improve user experience and also encourage Persian language businesses to grow in the tourism sector.

\section{Methodology}\label{sec3}
This section outlines the dataset, preprocessing steps, model architecture, training procedure, and evaluation strategy used in this study. This research presents a multi-stage model for ABSA, for the Aspect Category Detection (ACD) subtask, which is aimed at recognizing and categorizing key aspects of the Jabama platform (www.jabama.com) Persian user reviews. Our method includes gathering and preparing Persian data, creating the system architecture \citep{AfsheenMaroof2024}, training the model, and performing a comprehensive evaluation.

\begin{figure}[h]
    \centering
    \includegraphics[width=1\linewidth]{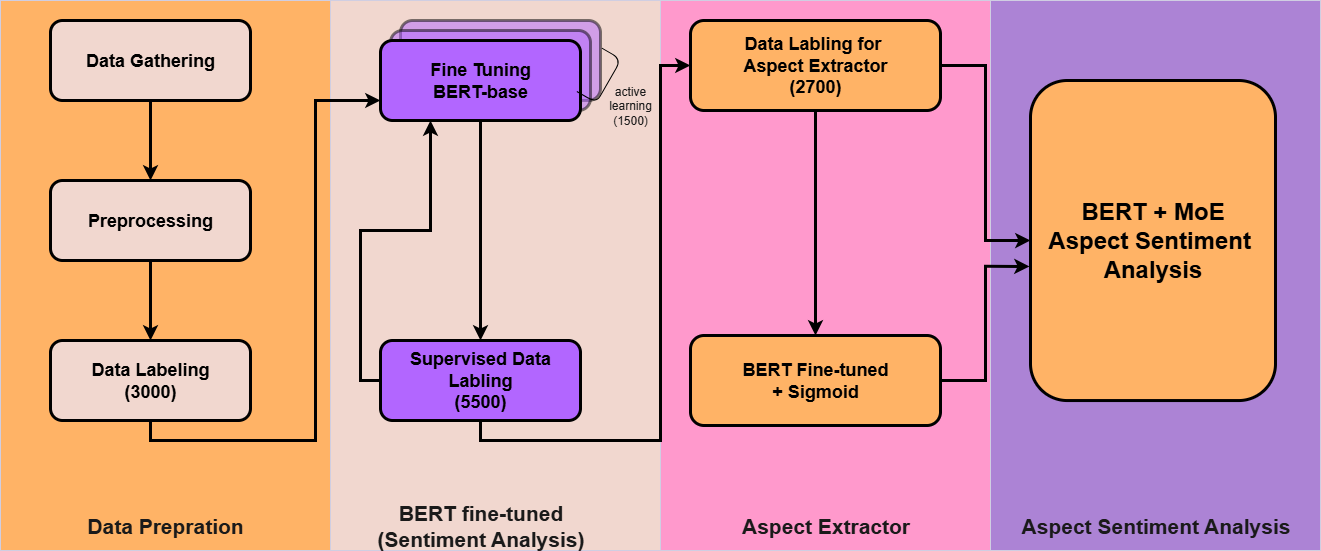}
    \caption{Workflow of data collection, preprocessing, model training and evaluation.}
    \label{fig1}
\end{figure}

\paragraph{Dataset Collection and Preprocessing.}
A dataset of 72,238 user reviews was collected from Jabama, a leading Iranian tourism platform that serves over seven million users and 18,000 hosts across 769 cities. Because of the irregular characters, inconsistent half-spaces, varying forms of orthography, and general differences regarding the spelling of words in the language, preprocessing became necessary \citep{Ghafouri2023,Rajabi2021,Nandwani2021}. Standardizing characters as well as removing emojis, ensuring uniform application of half spaces, correcting over a hundred common spelling errors, unifying vocabularies, splitting concatenated words, and removing irrelevant spam were some tasks in the preprocessing pipeline.

\begin{figure}[h]
\centering
\includegraphics[width=0.6\textwidth]{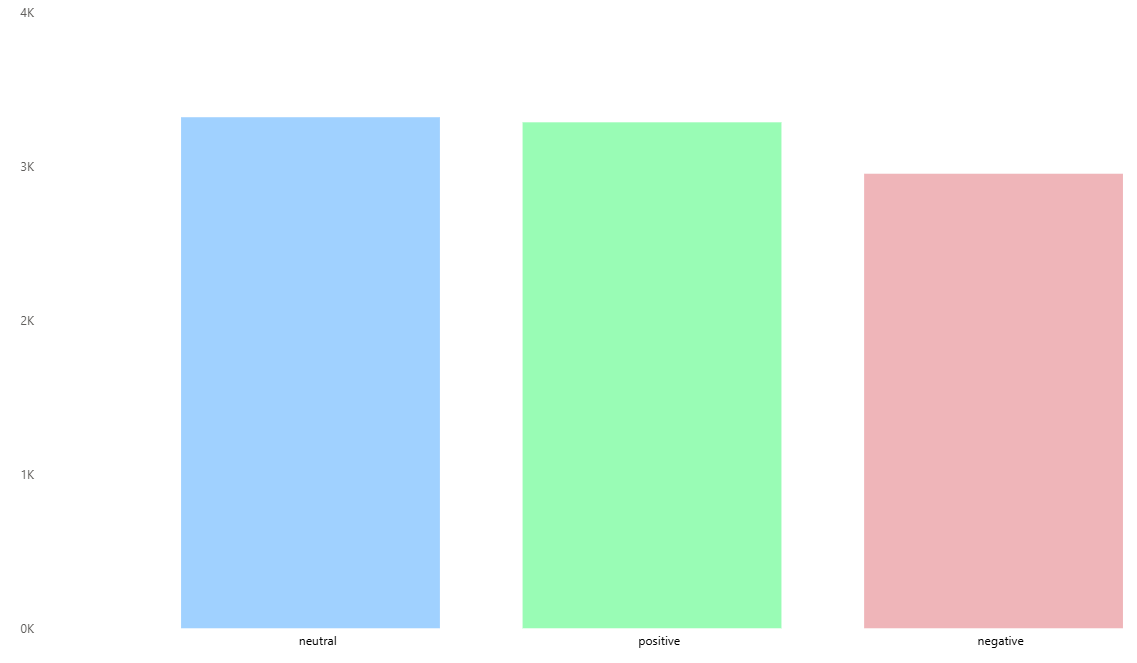}
\caption{Distribution of 9,558 data points used for training the BERT base model.}
\label{fig2}
\end{figure}

58,473 high-quality reviews were kept. Each review was assigned a sentiment polarity and classified into six major categories: host, price, location, amenities, cleanliness, and connectivity. The categories were inspired by sentiment ABSA schemas designed for Persian \citep{MorenoOrtiz2019,Ataei2019,Afzaal2019}, ensuring precise and high-quality annotations for subsequent modeling tasks. Similar domain-specific annotation frameworks have been validated for tourism reviews to improve inter-annotator agreement and schema reliability \citep{MorenoOrtiz2019}.

\paragraph{Model Development.}
The proposed model was developed in three stages (Figure~\ref{fig1}):

\begin{enumerate}
\item \textbf{Basic Sentiment Analysis}\\
For fundamental sentiment analysis, we adjusted a BERT Base model \citep{Devlin2019} to categorize review sentiments into positive, negative, and neutral classes. To address limited data, we used a semi-supervised active learning technique, whereby the model was repeatedly used on the unlabeled data. To begin, we evaluated and manually incorporated 1,800 high-confidence predictions into the training data. Subsequently, we integrated the rest of the high-confidence predictions automatically, resulting in the final labeled data set consisting of 9,558 samples. The modified model achieved an F1 score of 93.3\% (with a learning rate of $2 \times 10^{-5}$, batch size of 32, and 4 epochs).
\begin{table}[h]
\caption{This section shows a comparison of Persian sentiment analysis models. Considering the different datasets utilized, the comparison is indirect. It is only a general guide to compare performance\citep{DeepSentiPers2020,Farahani2021}}
\label{tab1}
\centering
\begin{tabular}{lcc}
\toprule
Model & Dataset & F1-score (\%) \\
\midrule
BERT fine-tuned & Jabama & 93.3\% \\
ParsBERT v2 & SentiPers (Binary Class) & 92.42\% \\
DeepSentiPers & SentiPers (Binary Class) & 91.98\% \\
\botrule
\end{tabular}
\end{table}
\begin{figure*}[t]
    \centering
    \includegraphics[width=0.32\textwidth]{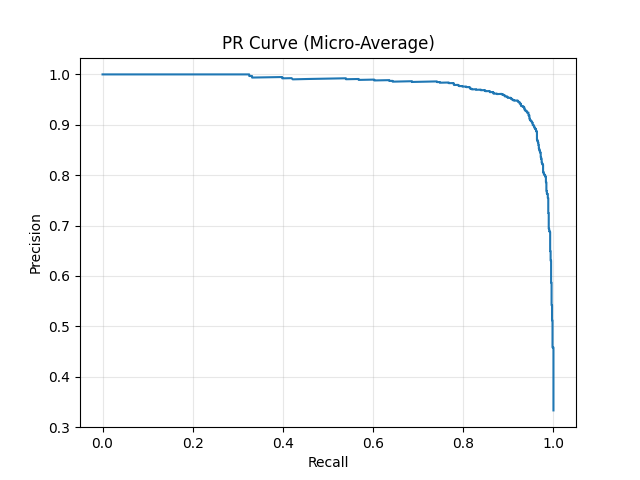}
    \includegraphics[width=0.32\textwidth]{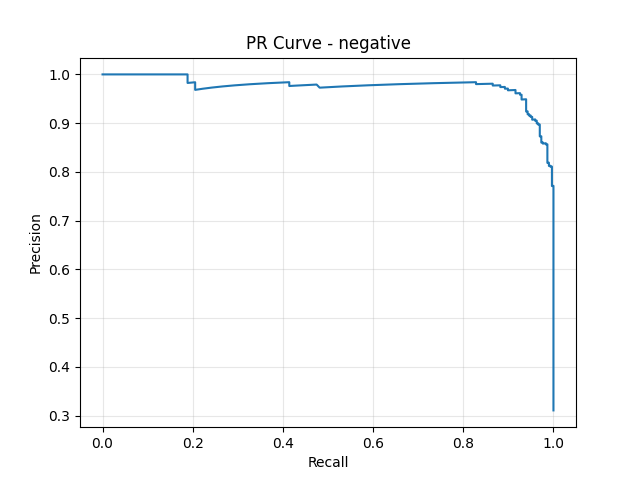}
    \includegraphics[width=0.32\textwidth]{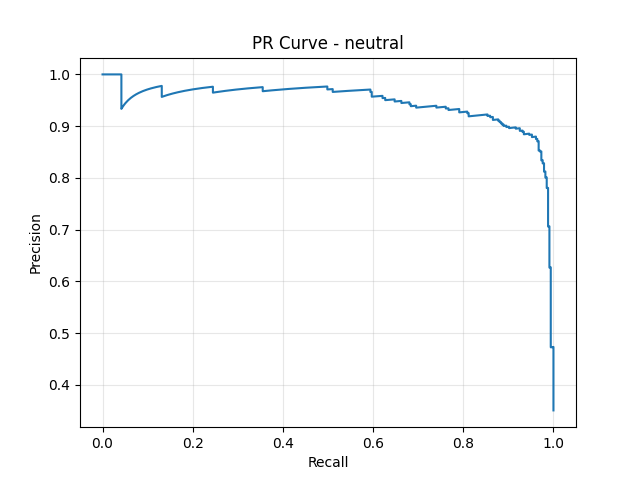} \\
    \includegraphics[width=0.32\textwidth]{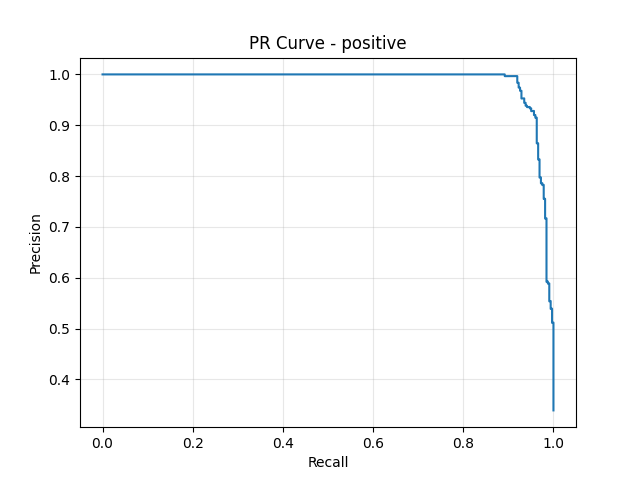}
    \includegraphics[width=0.32\textwidth]{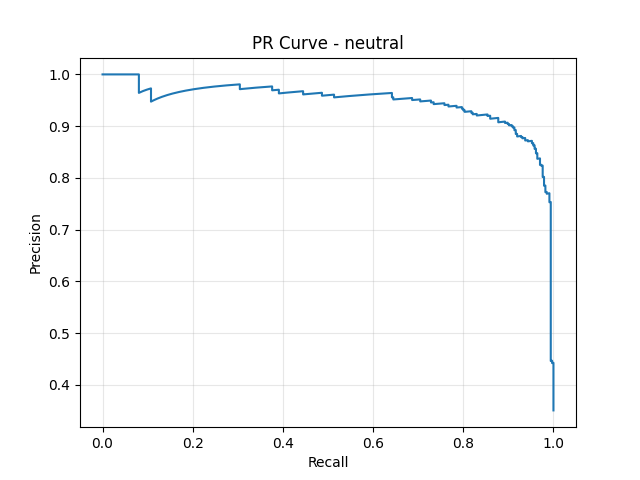}
    \includegraphics[width=0.32\textwidth]{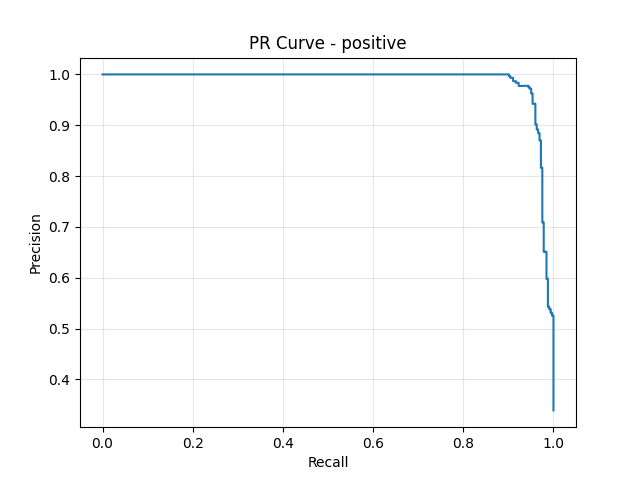}
    
    \caption{Precision--Recall curves for the baseline BERT model. Overall classification performance is summarized with the micro-average, which is located in the top left corner. Per-class curves show the results for the negative, neutral, and positive sentiment classes. The neutral class exhibits greater fluctuations than the other classes due to imbalance. In contrast, the positive and negative classes display more stable behavior.}
    \label{fig3}
\end{figure*}

\item \textbf{Aspect Category Detection (ACD)}\\
A modified BERT encoder with a sigmoid activation function (Figure~\ref{fig5}) was trained to identify six aspects: host, price, location, amenities, cleanliness, and connectivity. While achieving a weighted F1 score of 89.69\% during training, the following hyperparameters were configured: (1.7e-5 learning rate, 8 batch size and 4 epochs). Two domain-aware annotators maintained consistency and clarity throughout the annotation process, as for every aspect, a set of predefined semantic definitions was provided. The six aspect categories are host, location, amenities, connectivity, cleanliness, and price. The host aspect accounts for statements regarding the demeanor of the owner or the reception staff. Location refers to comments regarding geographic accessibility, views, and positioning in general. Amenities cover facilities within the dwelling, including the kitchen, swimming pool, systems for heat regulation (packaged units), ventilation, and security. Comments regarding the access to the internet or the mobile signal are addressed in connectivity (e.g., ``no signal,'' ``weak Wi-Fi''). The cleanliness aspect refers to comments related to the hygiene and tidiness of the rooms. Last, price refers to comments about value for money, cost, and the fairness of pricing. Reviews could be tagged with multiple annotations. A single review might comprise multiple category labels. Implicit sentiment was shown in phrases like `wish it was cleaner.' Reviews that didn't show the needed points were removed to keep things clear and good. Tabel~\ref{tab4} in the results section shows how the aspect labels are spread. Of the attributes, cleanliness and amenities were noted as two of the more prominent. This observation directly reflects the distribution shown in Figure~\ref{fig4}. This careful labeling helped train the multi-label classifier and made the ACD stage more reliable.

\begin{figure}[h]
\centering
\includegraphics[width=0.5\textwidth]{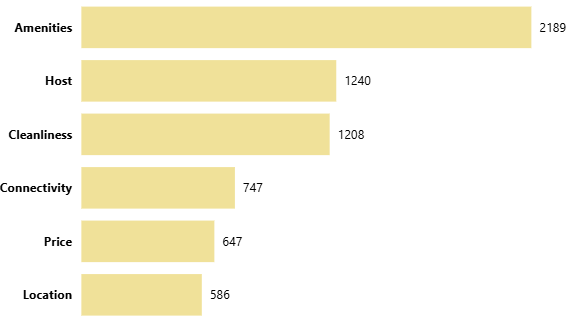}
\caption{The distribution of labels for aspects.}
\label{fig4}
\end{figure}

\begin{figure}[h]
\centering
\includegraphics[width=0.5\textwidth]{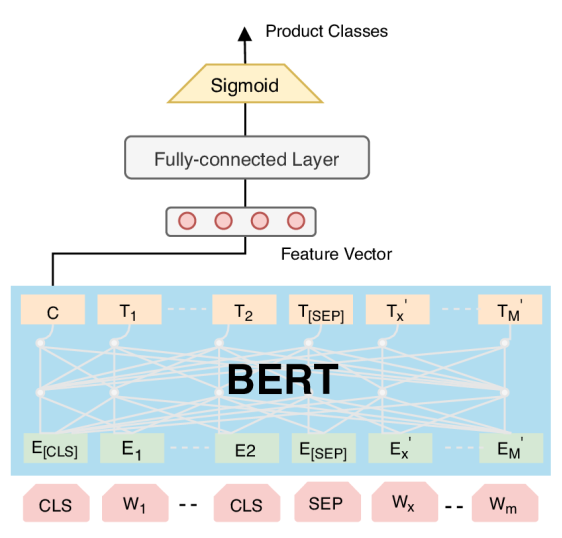}
\caption{Architecture of the BERT-based model used for Aspect Category Detection (ACD). A sigmoid activation function is applied to the output layer to allow multi-label classification across six aspect categories: host, price, location, amenities, cleanliness, and connectivity.}
\label{fig5}
\end{figure}

\item \textbf{Aspect-Based Sentiment with a hybrid BERT model}\\
To address the aspect-based sentiment classification (ABSA) task, we designed a Mixture-of-Experts (MoE) architecture atop a fine-tuned BERT model. The intuition behind this design is that different experts can specialize in different sentiment patterns across aspects such as cleanliness, price, location, and others, while a learned gating mechanism determines the contribution of each expert for a given input. Our MoE model consists of a BERT base encoder (bert-base-multilingual-cased) fine-tuned on domain-specific Airbnb review data \citep{Devlin2019}, with the [CLS] token embedding (dimension: 768) serving as input to six feed-forward neural network experts. Each expert comprises a linear layer (768→256), a ReLU activation, and a second linear layer (256→3) for the three sentiment classes (positive, neutral, negative). A gating network, receiving the aspect term embedding, outputs a softmax distribution over the experts, producing weights for a batch-wise einsum operation to aggregate expert outputs. This specialized routing follows the Mixture-of-Experts paradigm \citep{Shazeer2017} and incorporates recent advances in top-k routing \citep{Zeng2024}.

\begin{table}[h]
\caption{Performance Comparison of ABSA Models.}
\label{tab2}
\centering
\begin{tabular}{lcc}
\toprule
Model & F1-score (\%) \\
\midrule
Standalone BERT & 89.25 \\
BERT+MoE & 89.43 \\
BERT+MoE+LoRA(Hard gate) & 85.7 \\
BERT+MoE+LoRA & 85.7 \\
\botrule
\end{tabular}
\end{table}

\begin{table}[h]
\caption{Performance and Routing Analysis of MoE Variants.}
\label{tab3}
\centering
\begin{tabular}{lcc}
\toprule
Variant & F1-score (\%) & COV² \\
\midrule
BERT+MoE (Baseline) & 89.43 & 1.5856 \\
BERT+MoE + Aux Loss v1 & 93.03 & 2.1406 \\
BERT+MoE + Aux Loss v2 (MSE) & 93.36 & 2.0900\\
\botrule
\end{tabular}
\end{table}

This architecture achieved an overall F1-score of 90.6\% (learning rate = 1.8552 $\times 10^{-5}$, batch size = 8, epochs = 3), which is greater than the scores of BERT and advanced hybrid BERT model (BERT+MoE+LoRA). Compared to other models, the hybrid expert-enhanced BERT has about 164 million parameters and can balance complexity and performance. We used two MoE approaches, the first being a hard mapping variant where experts were fixed and assigned to specific aspects (F1 = 87.23\%), and the second being a dynamic routing variant where the gate leans into weighted aspect embeddings for the best score (F1 = 90.80\%). The dynamic routing facilitated expert specialization, as seen in the expert weight heatmap (Figure~\ref{fig6}). Other than the F1-score, which we weighted primarily because of imbalanced classes, the model used categorical cross-entropy loss. To better fine-tune hyperparameters, the dataset was divided into 80\% training and 10\% each for validation and testing. Future work can consider the fusion of sentence-aspect in the gate for greater interpretability, as well as attention visualization.

\end{enumerate}

\paragraph{Loss Formulations and Routing Mechanisms.}
Our model is trained by minimizing the standard categorical cross-entropy (CCE) loss:
\begin{equation}
\mathcal{L}_{\mathrm{CE}} = -\frac{1}{N} \sum_{i=1}^{N} \sum_{c=1}^{C} y_{i,c} \log(\hat{y}_{i,c}),
\end{equation}
where $N$ is the batch size, $C{=}3$ is the number of sentiment classes (positive, neutral, negative), $y_{i,c}$ is the ground-truth one-hot label, and $\hat{y}_{i,c}$ is the predicted probability from the softmax layer.

To counteract routing collapse, we incorporate an auxiliary importance loss inspired by GShard~\citep{Lepikhin2020} and later adopted in Switch Transformer~\citep{Fedus2022}. The complete training objective is
\begin{equation}
\mathcal{L} = \mathcal{L}_{\mathrm{CE}} + \lambda_{\mathrm{aux}} \mathcal{L}_{\mathrm{aux}} + \mathcal{L}_{\mathrm{mse}},
\end{equation}
with $\lambda_{\mathrm{aux}} = 0.011822$. Full mathematical formulations of all three loss terms are provided in Appendix~\ref{app:loss_routing}.

\begin{figure}[!h]
    \centering
    \includegraphics[width=0.5\textwidth]{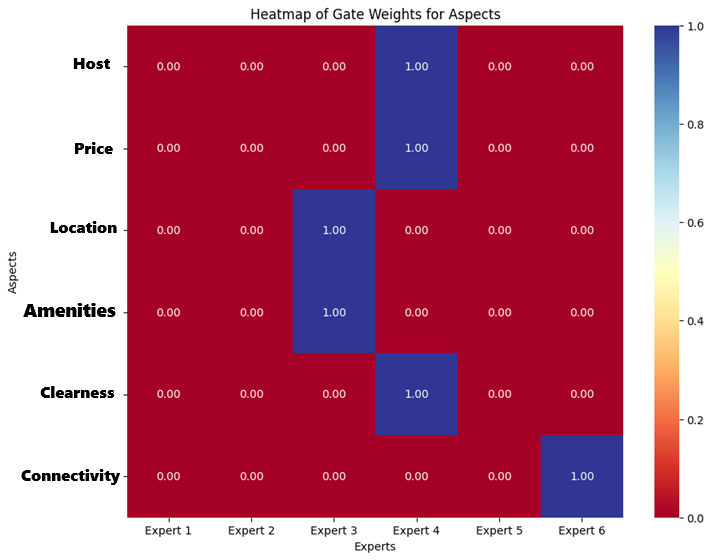}
    \caption{Heatmap of gate-assigned expert weights across aspect types (before applying rectification techniques), illustrating emergent specialization.}
    \label{fig6}
\end{figure}

\begin{figure}[h]
\centering
\includegraphics[width=0.5\textwidth]{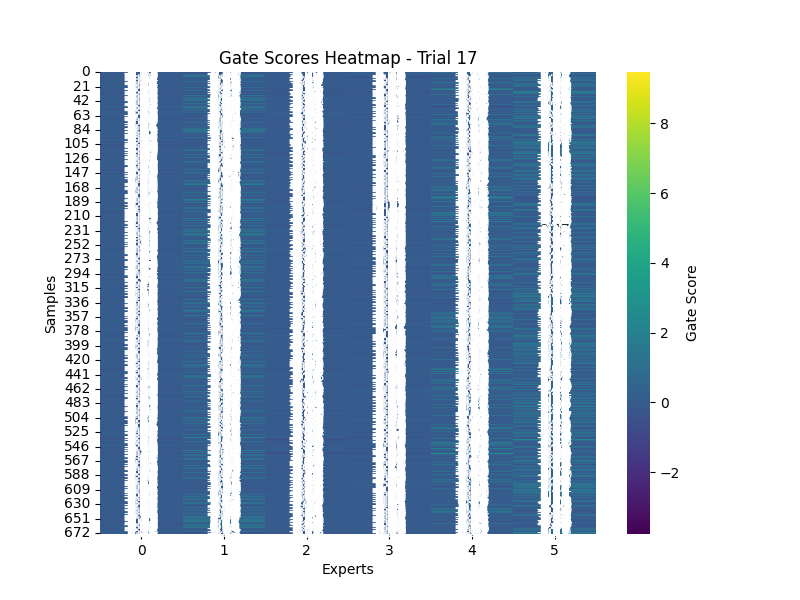}
\caption{Heatmap of gate-assigned expert weights across aspect types, illustrating improved specialization after Top-K routing implementation.}
\label{fig7}
\end{figure}

\paragraph{Enhanced Expert Utilization.}
We employed a Top-K routing mechanism ($K{=}3$) with a capacity factor of 1.8, combined with two rectification techniques to mitigate routing collapse. 
Intra-GPU Rectification (IR) reassigns dropped tokens (due to capacity overflow) to the highest-scoring local expert on the same GPU rather than discarding them. 
Fill-in Rectification (FR) fills padding positions in under-utilized experts with the $(k{+}1)$-th highest-scoring token candidates. During training, noisy Top-K gating was applied by adding Gumbel-distributed noise (scaled by 0.098323) to the gate logits. These modifications reduced the squared coefficient of variation (COV$^2$) of expert utilization from 1.5856 (baseline softmax routing) to 0.0109, achieving near-uniform load across all six experts (ideal balanced activity $\approx [1.0, 1.0, 1.0, 1.0, 1.0, 1.0]$). 
Specialization patterns before and after applying the rectification techniques are shown in Figures~\ref{fig6} and~\ref{fig7}, respectively. Full mathematical details of IR and FR, straight-through gradient handling, and complete pseudocode are provided in Appendix~\ref{app:loss_routing}.

\begin{figure*}[t]
    \centering
    \includegraphics[width=0.24\textwidth]{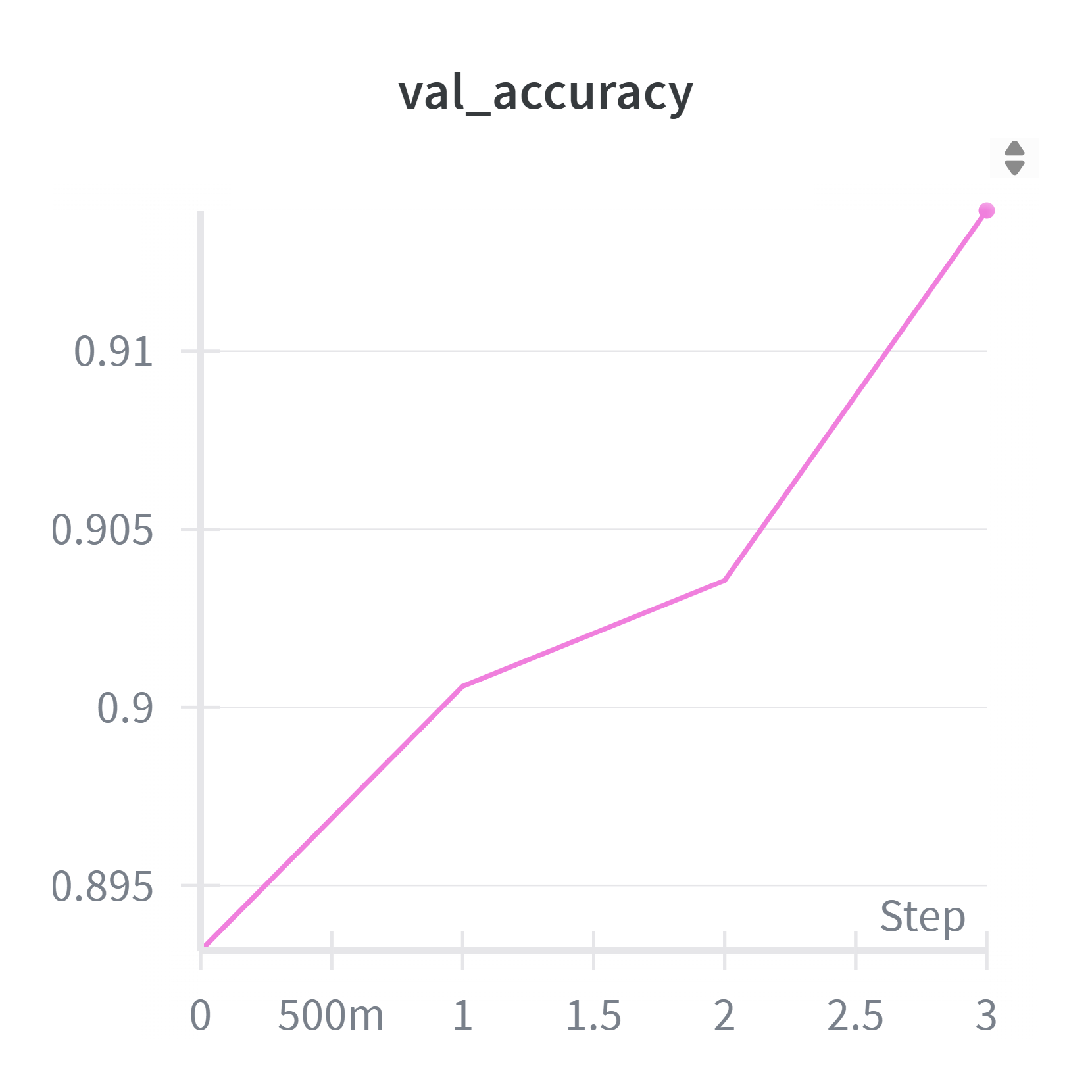}
    \includegraphics[width=0.24\textwidth]{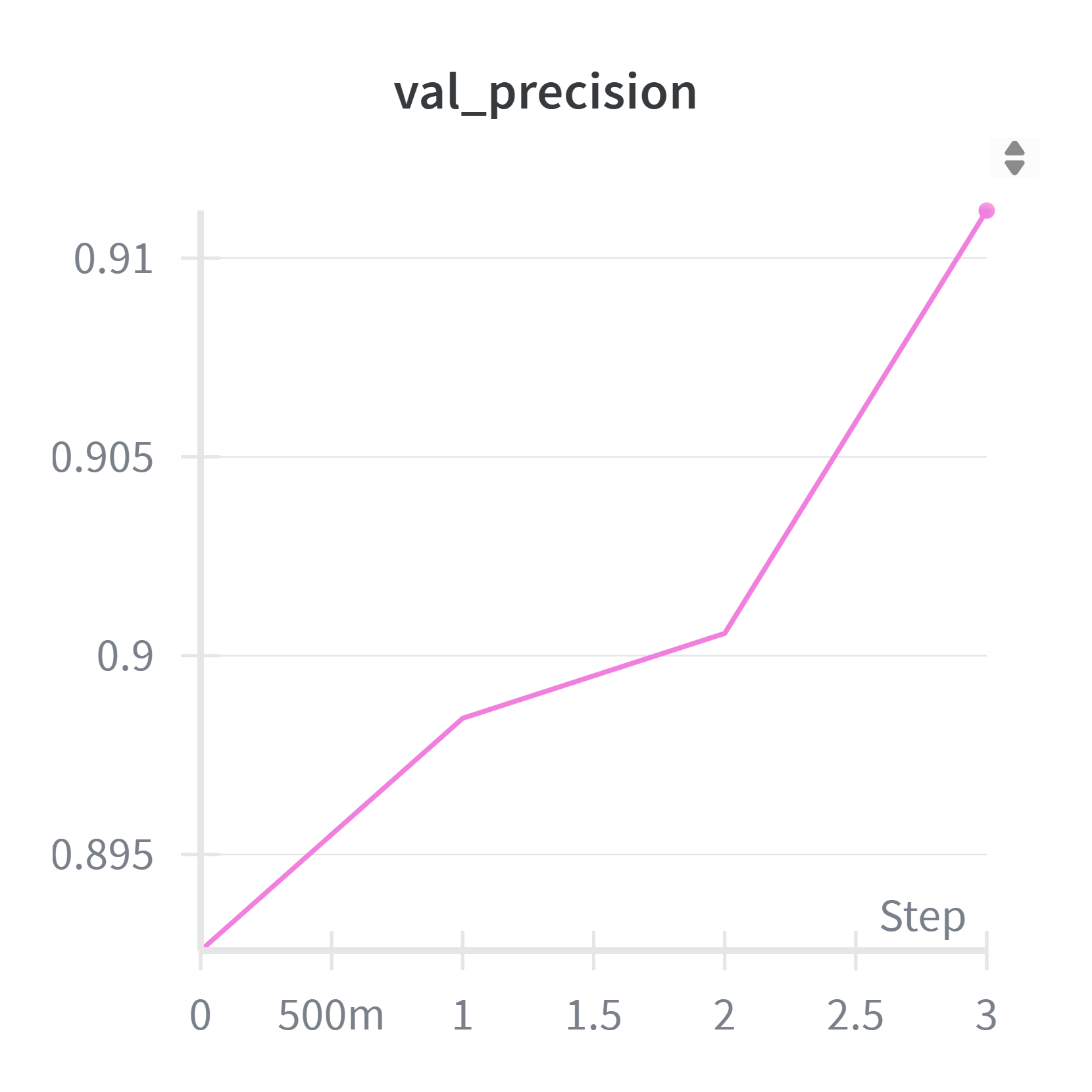}
    \includegraphics[width=0.24\textwidth]{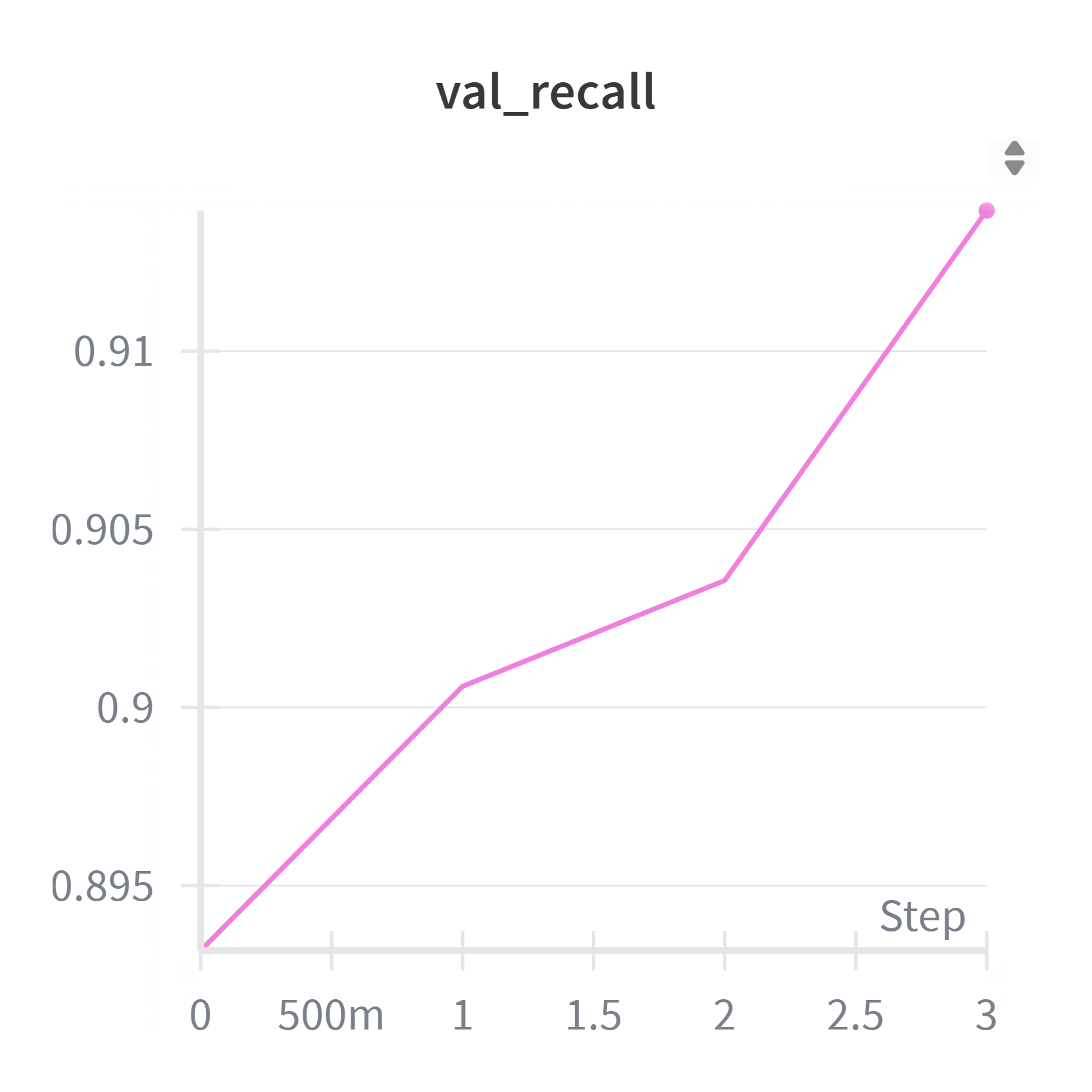}
    \includegraphics[width=0.24\textwidth]{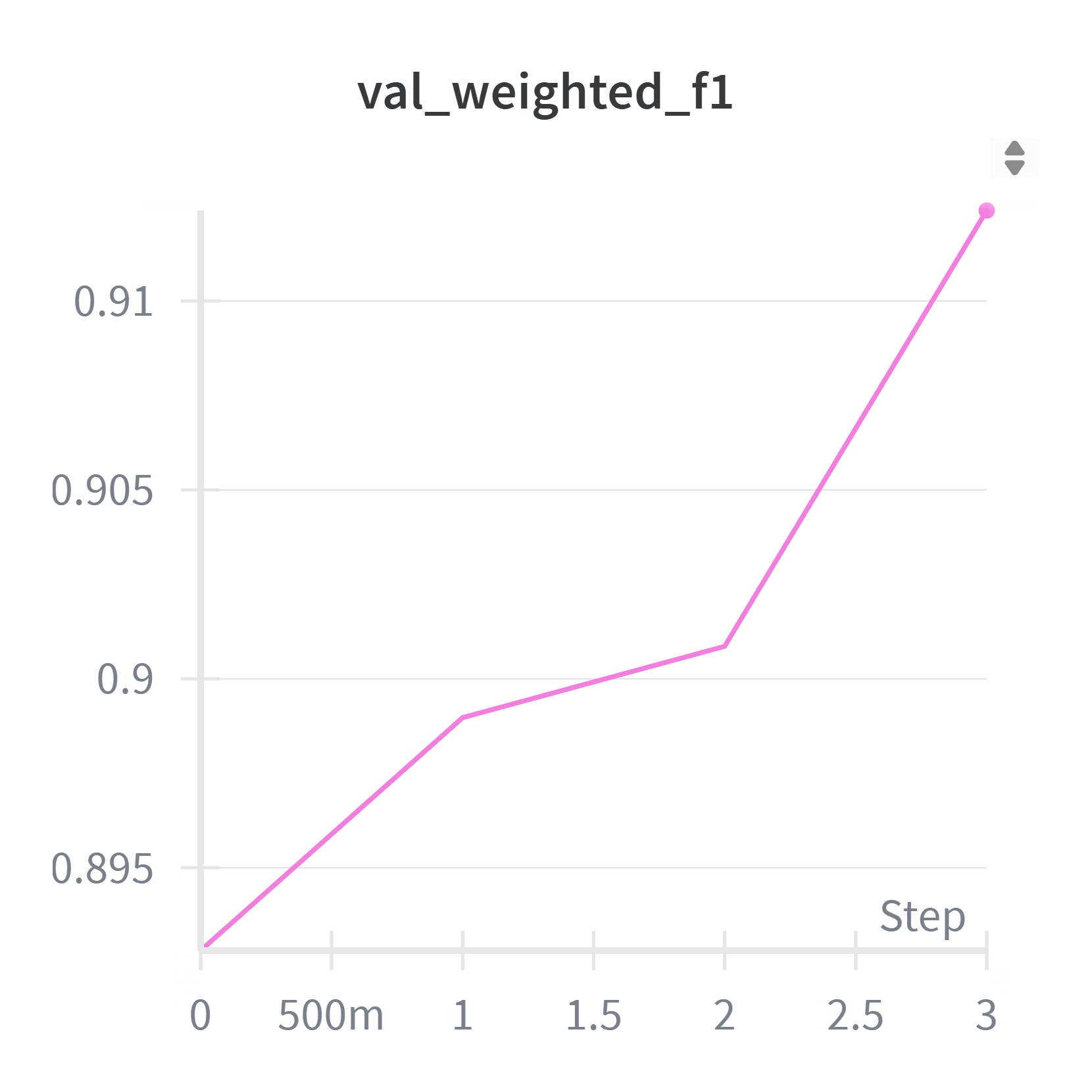} \\
    
    \caption{Validation performance metrics for the modified Top-K routing the hybrid expert-enhanced BERT model. From left to right: (a) Validation accuracy, showing stable convergence; (b) Validation precision, highlighting improved precision across sentiment classes; (c) Validation recall, illustrating robust performance despite class imbalance; (d) Weighted F1-score, reflecting superior overall performance.}
    \label{fig8}
\end{figure*}

The Top-K routing approach successfully resolved the routing collapse issue, enabling full and fair utilization of all experts in the MoE architecture. This significantly improved the model's generalization on unseen samples, as shown with the validation metrics in Figure~\ref{fig8}.

The model’s success in handling the inherent class imbalance in the dataset is evident from the consistent and high validation accuracy, as well as recall and precision metrics across all sentiment categories. The class imbalance, particularly the dominance of the cleanliness and amenities aspects, is often difficult to manage with standard classifiers. The weighted F1 score, which accounts for class distribution in its computation, provides further evidence of the proposed model’s superiority, showing a robust precision-recall trade-off that ensures balanced performance across both majority and minority classes.

\paragraph{Implementation Algorithm.}
The complete pseudocode for the MoE-based ABSA algorithm, including the Top-K routing with rectification mechanisms, is provided in Appendix~\ref{app:pseudocode}.
Persian language, stem from the language’s morphological and syntactic complexity and differing standards of orthography. Other issues with available labeled data and learning reliable representations exacerbates the problem, making the development of Persian NLP systems that perform at the level of English NLP systems almost impossible. In fact, the Top-K routed BERT with Mixture of Experts models integration you proposed could perform the functions of the BERT in outperforming language and NLP processing tasks in English and allow Persian NLP systems of comparable performance to be developed. The use of adaptive noise scaling, designed to temporally regulate the tradeoff between exploitation and exploration, is likely to be the most important for controlling training noise. The proposed technique may contribute to the quest for expert diversification, without over-specialization, to improve robustness for which more abstract objectives are often proposed. Plans that seek to enhance the self-organizing nature of the model, such as dynamically controlling capacity and expanding the model based on the complexity of the input, will increase the model’s robustness and computational efficiency for cross-domain tasks. The collaboration of active attention control in models, as proposed, will enhance model interpretability. Describing how the gating system selects various experts for the various segments of the task can elucidate how the model arrives at its decisions, thus offering some degree of explainability for the model itself. This model explainability is beneficial to both the user and the researcher. All of the aforementioned can increase the applicability of Top-K routing the hybrid expert-enhanced BERT frameworks for actual tourism platforms and also for numerous low-resource languages.

\begin{figure*}[t]
    \centering
    \includegraphics[width=0.24\linewidth]{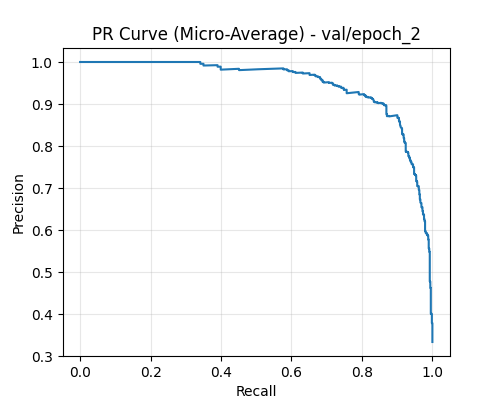}
    \includegraphics[width=0.24\linewidth]{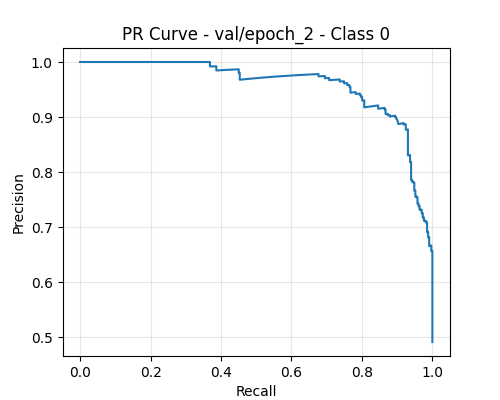}
    \includegraphics[width=0.24\linewidth]{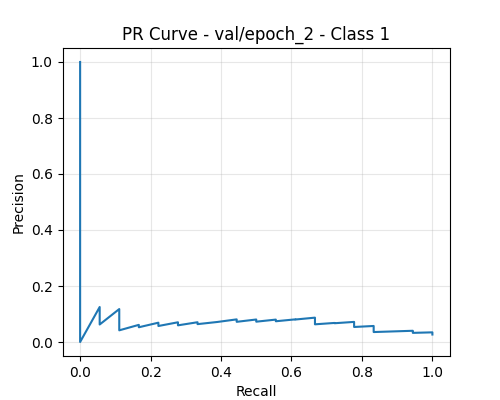}
    \includegraphics[width=0.24\linewidth]{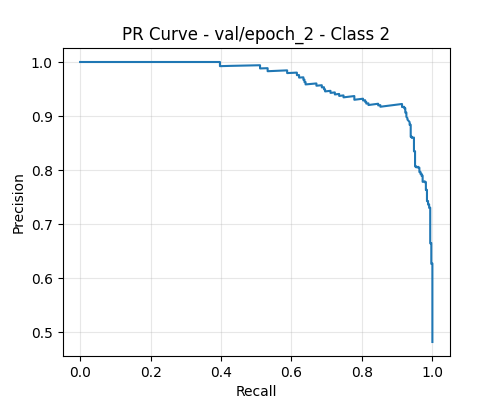}
\caption{The precision-recall curves were generated to assess the performance of the hybrid expert-enhanced BERT model on the validation set across each sentiment class (Class 0 - negative, Class 1 - neutral, Class 2 - positive). Among the three, Class 1 (neutral) curves stand out as the most irregular with pronounced fluctuations, while the curves about to the negative and positive classes exhibit greater stability. The micro-average curve indicates the model's overall performance, while the class-average curves delineate performance regarding each specific sentiment.}
\label{fig9}
\end{figure*}
\begin{figure*}[t]
    \centering
    \includegraphics[width=0.24\linewidth]{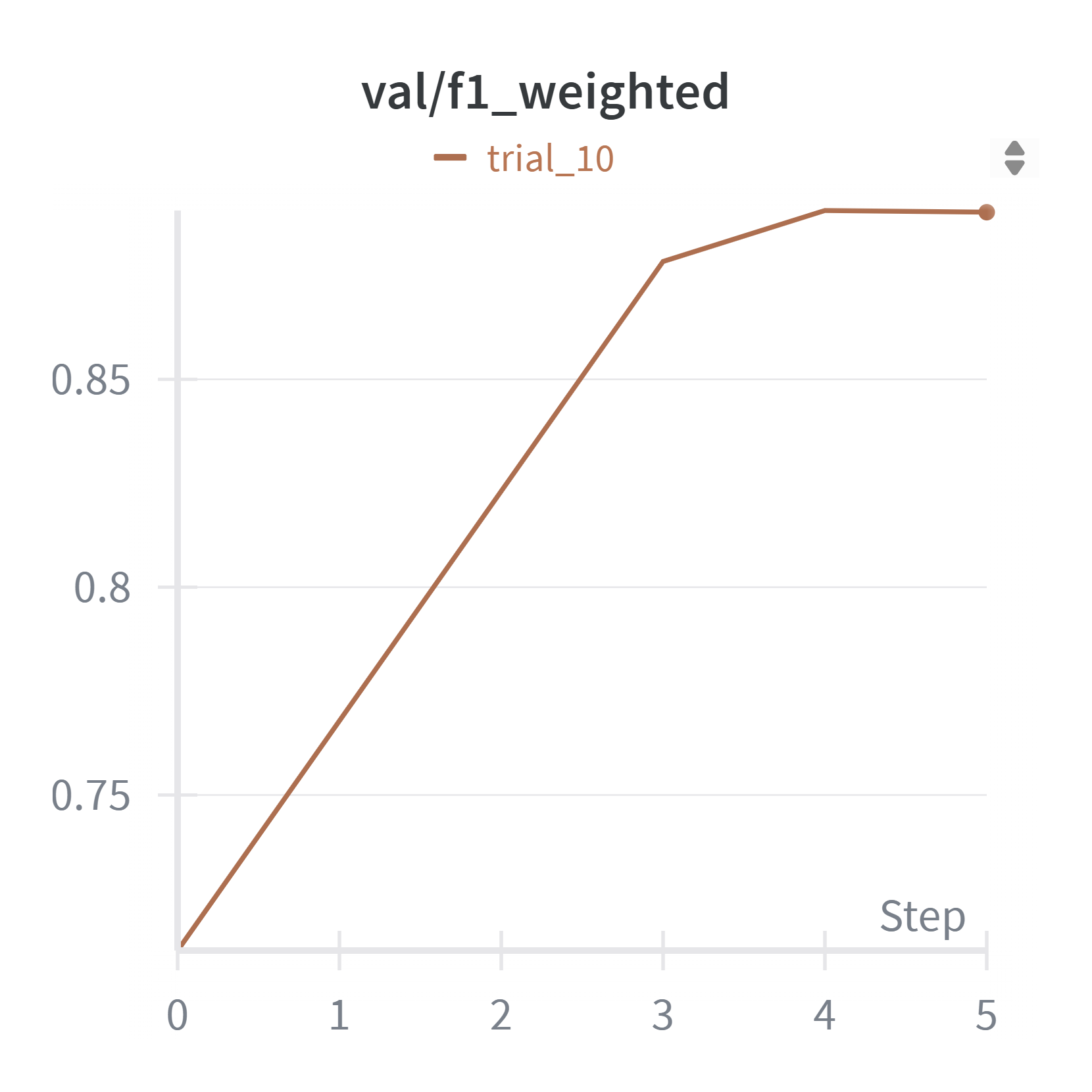}
    \includegraphics[width=0.24\linewidth]{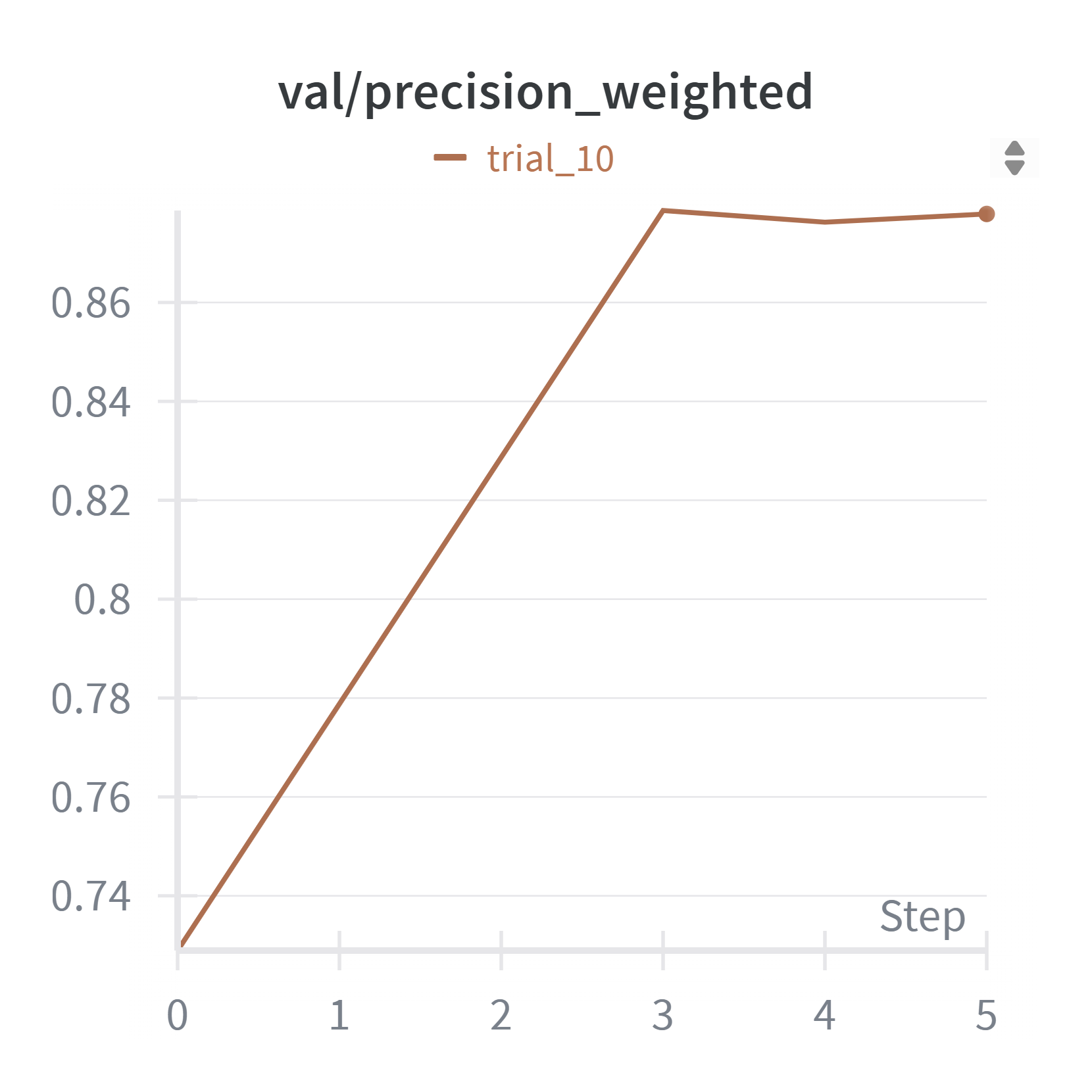}
    \includegraphics[width=0.24\linewidth]{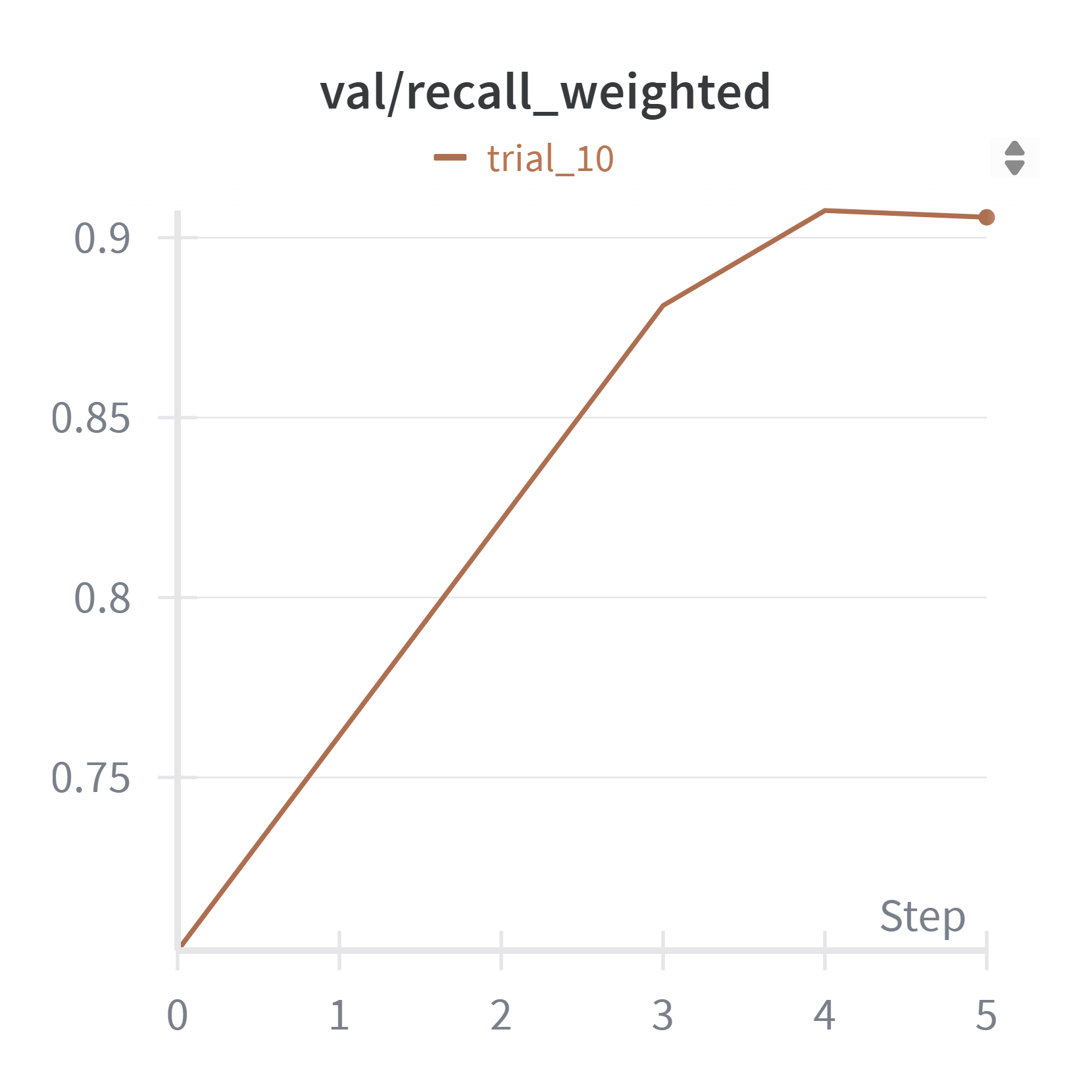}
    \includegraphics[width=0.24\linewidth]{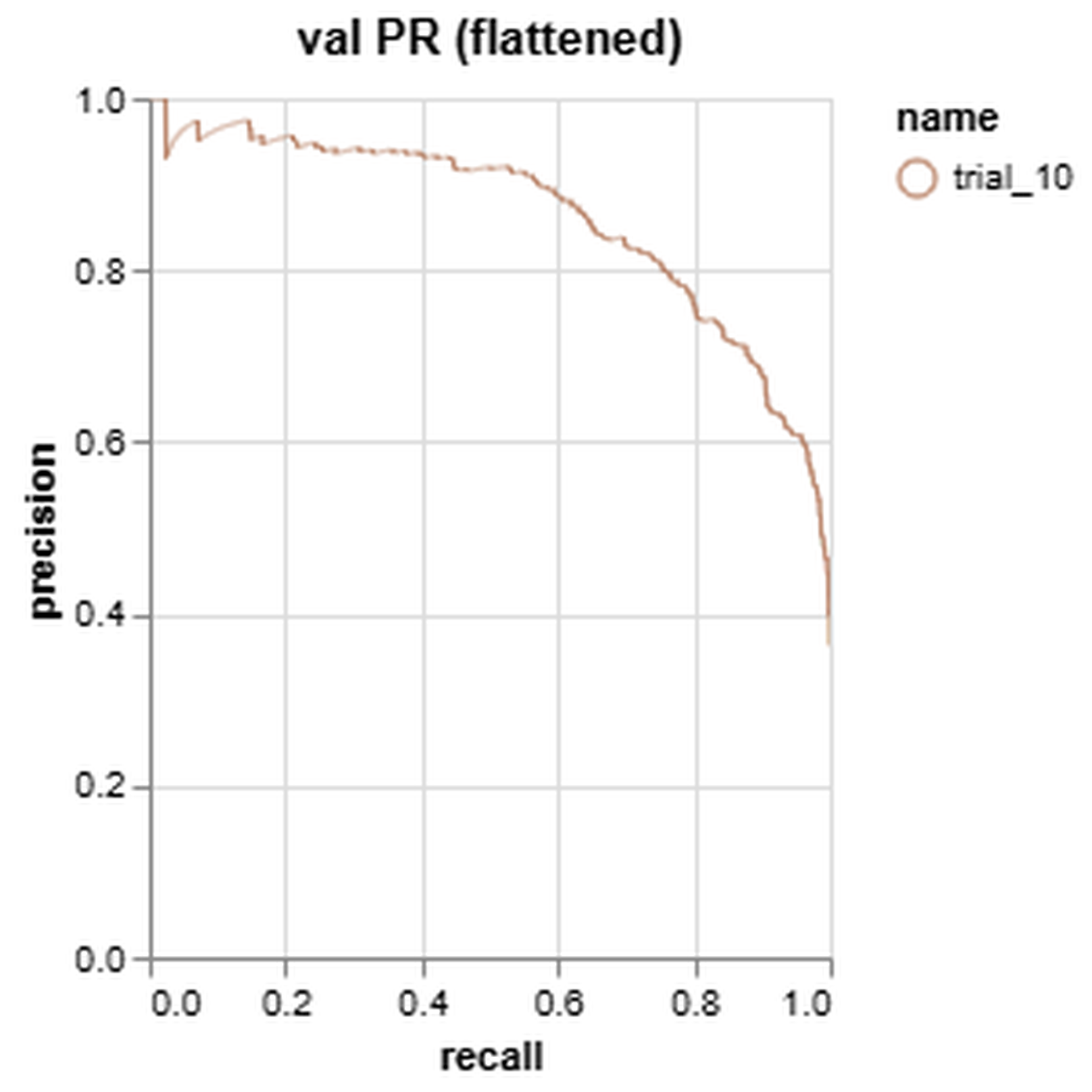}
    
\caption{Results of the Validation Aspect Category Detection (ACD) Model. The weighted precision-recall curve indicates the adjustments in the precision-recall tradeoffs for all aspects according to class size. The improvement of the weighted F1-score during training shows a constant increase in overall model performance. The primary increase in weighted precision and recall further indicates the model’s capability in handling and processing the challenges presented by unbalanced aspect categories.}
\label{fig10}
\end{figure*}

\begin{figure*}[p]
\centering
\includegraphics[width=1\linewidth,height=0.8\textheight,keepaspectratio]{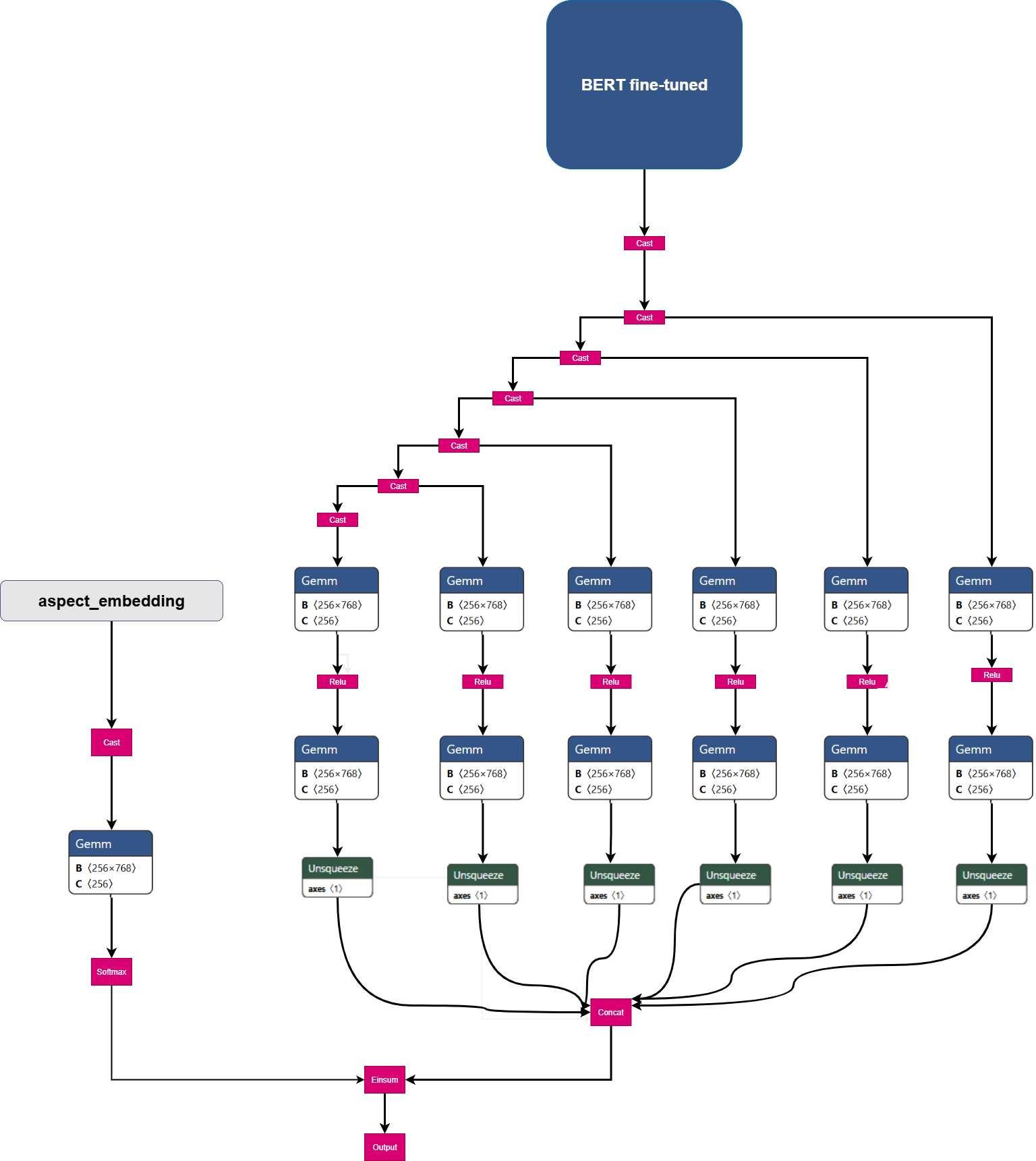}
\caption{Schematic of the the hybrid expert-enhanced BERT architecture, showing sentence embedding flow into experts and soft routing based on aspect embeddings.}
\label{fig:fig11}
\end{figure*}
\section{Results}\label{sec4}

The outcomes related to the proposed models during all three stages of development can be found in Table~\ref{tab1}. The sentiment analysis BERT model completed the Basic Sentiment Analysis stage, having scored 93.3\% weighted F1. This shows the model successfully classified the reviews as positive, negative, or neutral with consistent high accuracy. In the Aspect Category Detection (ACD) stage, the BERT encoder with a sigmoid activation function scored 88.0\% F1 score across the 6 predefined aspects, which were host, price, location, amenities, cleanliness and connectivity. The hybrid expert-enhanced BERT model achieved the highest performance in aspect-based sentiment analysis (ABSA) because it surpassed a 90.6\% weighted F1 score, which was also better than the scores of its standalone BERT (a weighted F1 score of 89.25\%) and advanced hybrid BERT model (BERT+MoE+LoRA) (a weighted F1 score of 85.7\%). This can be found in Table~\ref{tab2}. The precision--recall dynamics across distinct sentiment classes, as well as the overall microaverage performance, is evaluated in Figures~\ref{fig3}, \ref{fig9}, and \ref{fig10}. The analysis BERT model scored high on precision and recall stability in positive and negative classes, while the neutral class showed greater fluctuations in performance stability because of class imbalance. For the ACD model, the micro-averaged PR curve shows powerful performance across all aspects, and the macro metrics show that it also performs well on less frequent categories. The hybrid expert-enhanced model achieved an excellent trade-off between precision and recall across the various sentiment classes, and the dynamic routing mechanism was key to making consistent gains over the baseline. A major difficulty in this study was the imbalance in the distribution of aspect labels, as detailed in Table~\ref{tab4}. Cleanliness and amenities were mentioned more in reviews than connectivity and the host, which could lead the model to be biased. However, the BERT model used for baseline sentiment analysis pivoted around this using contextual embeddings to detect sentiment in a much more complex way, which achieved an impressive weighted F1 score of 93.3\%. The next hybrid expert-enhanced model architecture focused on this issue by using aspect-focused specialized experts and a gating mechanism that dynamically allocates weights to experts. The heatmap (Figure~\ref{fig6}) shows that some experts focused mainly on specific aspects. This made the model perform better on datasets with imbalances.

\begin{table}[h]
\caption{Distribution of aspect categories and sentiment labels in the dataset. Sentiment labels: Negative, Neutral, Positive.}
\label{tab4}
\centering
\begin{tabular}{lrrrr}
\hline
\textbf{Aspect} & \textbf{Total} & \textbf{Negative} & \textbf{Neutral} & \textbf{Positive} \\
\hline
Price                & 693 & 579 & 40 & 128 \\
Amenities            & 2202 & 1508 & 73 & 621 \\
Host                  & 1267 & 188 & 15 & 1064 \\
Location              & 608 & 187 & 10 & 411 \\
Cleanliness        & 1228 & 359 & 30 & 839 \\
Connectivity          & 749  & 580 & 40 & 129 \\
\hline
\end{tabular}
\label{tab:aspect_distribution}
\end{table}
                            
The results indicate that the proposed model could cope with the challenges of the Persian language, achieving excellent results with high F1-scores. This may indicate the continuation of NLP research on the Persian language and potentially other under-resourced languages. The architecture of the hybrid expert-enhanced model, with its modularity and dynamic routing, should be able to be generalized to other datasets, particularly within the tourism domain. Due to domain-specific attributes like host, price, and location, if enough labeled data is provided, the model can be used for other related platforms, such as international booking services. This model could help tourism platforms perform user feedback analysis in a more meaningful way. For example, service providers can detect what needs improvement, like cleanliness or internet access, and travelers can choose better based on detailed feedback. This helps users have a better experience and improves the connection between hosts and guests in Iran’s growing digital tourism sector.
Although the overall weighted F1 improvement appears modest (+1.35 percentage points over the dense BERT baseline), 
the proposed MoE architecture delivers two decisive advantages that strongly justify its added complexity:

\begin{itemize}
  \item \textbf{Energy efficiency}: A 39\% reduction in GPU power consumption compared to dense BERT (Figure~\ref{fig12}), directly supporting UN SDG 12 on responsible consumption and enabling cost-effective, sustainable deployment on tourism platforms in developing regions.
  \item \textbf{Training stability and scalability}: Near-elimination of routing collapse (COV$^2$ reduced from 1.5856 to 0.0109) ensures stable long-term training and straightforward horizontal scaling—critical limitations that have historically hindered practical adoption of MoE models in real-world, low-resource settings.
\end{itemize}

These benefits make the architecture particularly well-suited for applications where sustainability, operational cost, and reliable scaling are prioritized alongside predictive performance.

\clearpage                                 
\begingroup                                
\renewcommand{\topfraction}{0.95}         
\renewcommand{\bottomfraction}{0.95}       
\renewcommand{\textfraction}{0.05}         
\renewcommand{\floatpagefraction}{0.9}     
\setcounter{topnumber}{2}                  
\setcounter{bottomnumber}{2}
\setcounter{totalnumber}{4}

\begin{figure*}[t!]                        
    \centering
    \includegraphics[width=0.32\textwidth]{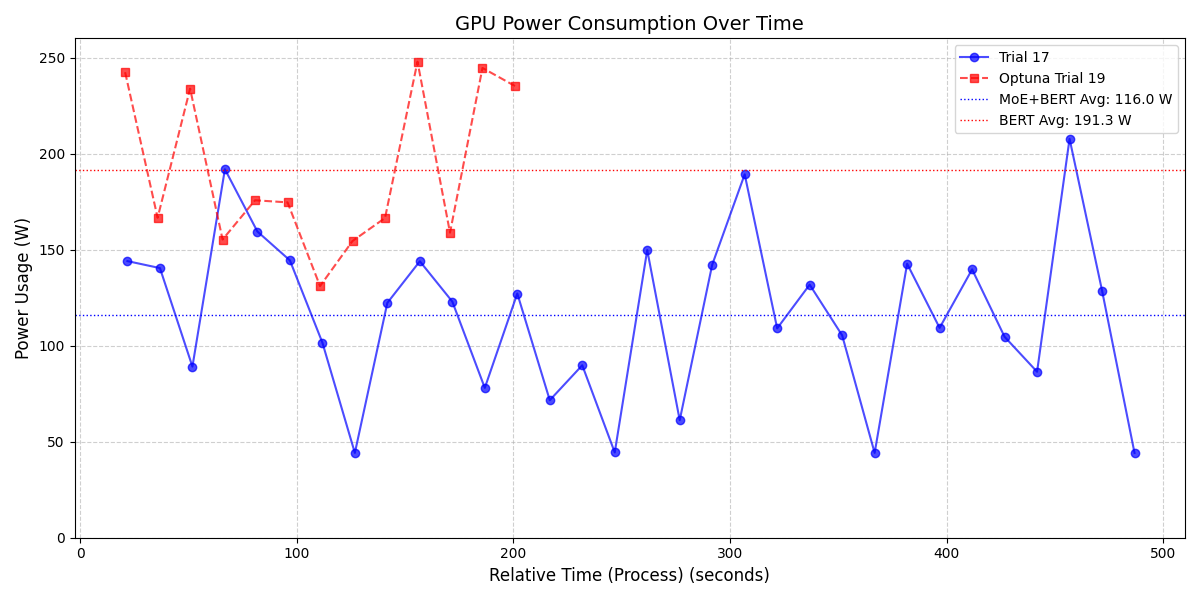}
    \includegraphics[width=0.32\textwidth]{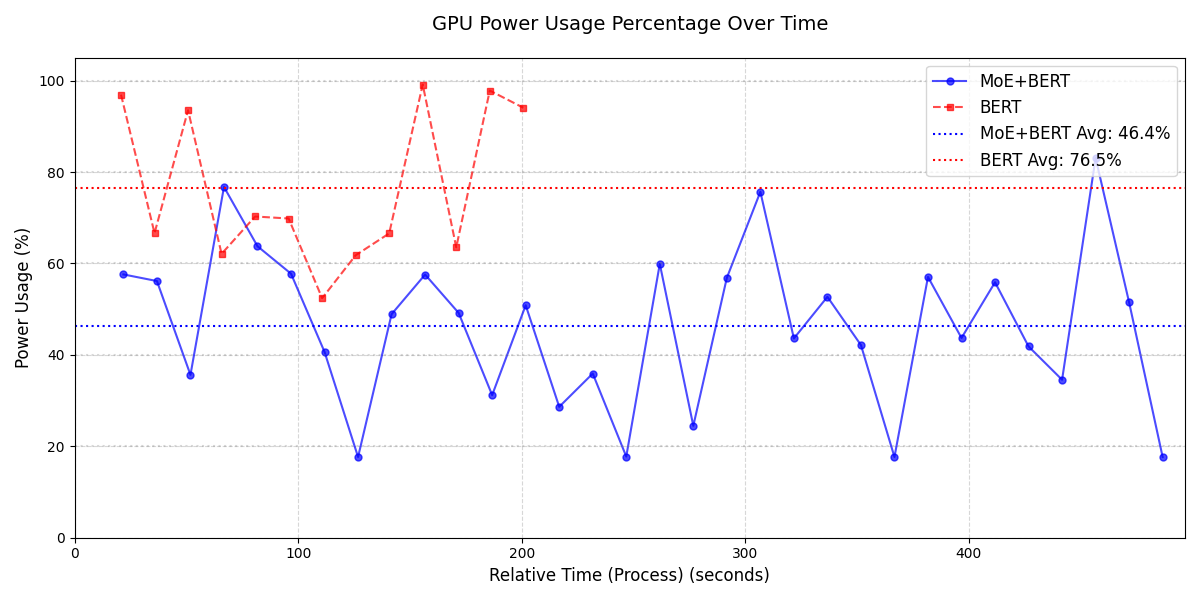}
    \includegraphics[width=0.32\textwidth]{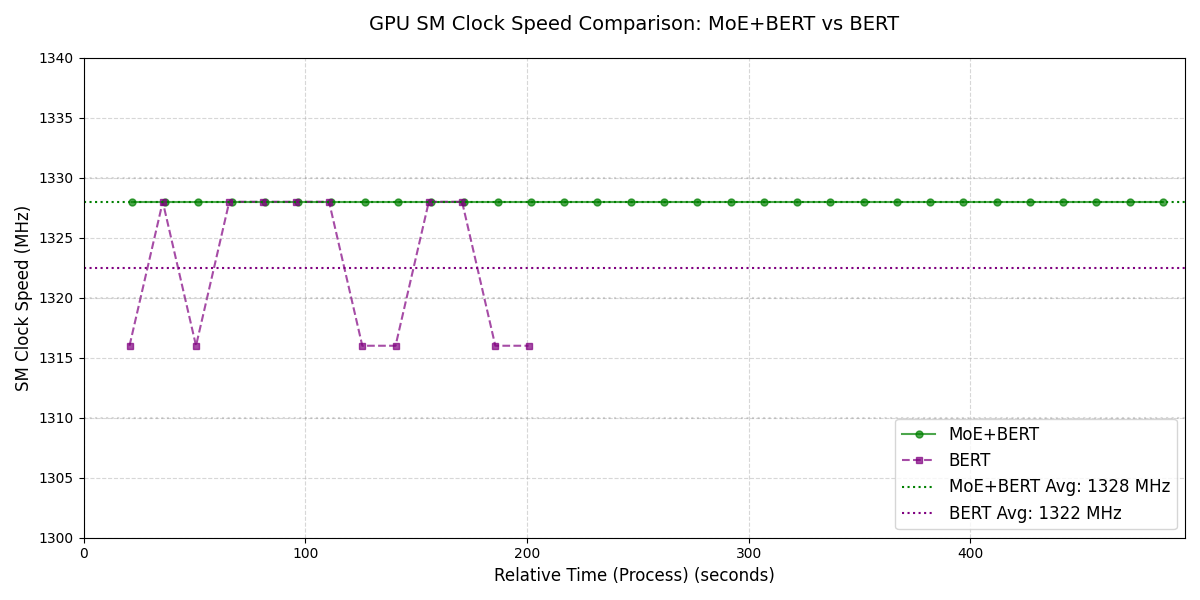}\\
    \vspace{0.2cm}
    \includegraphics[width=0.32\textwidth]{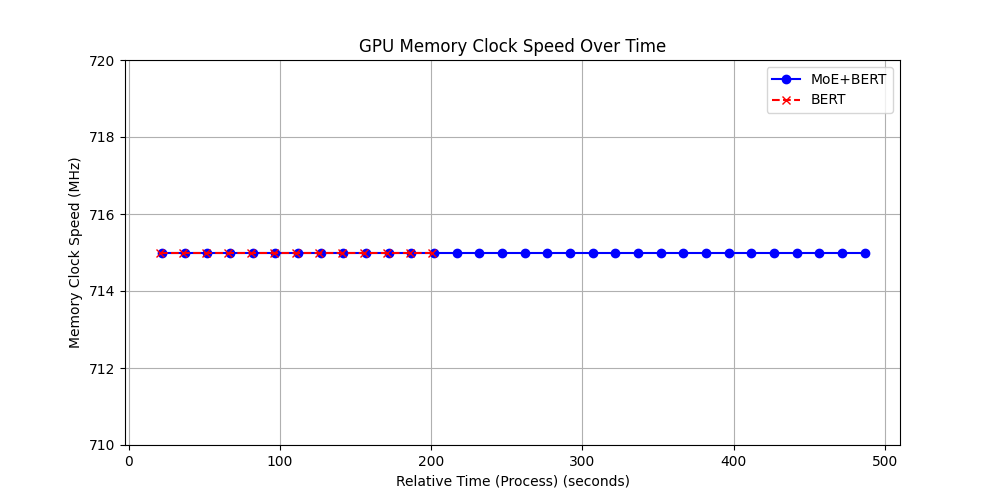}
    \includegraphics[width=0.32\textwidth]{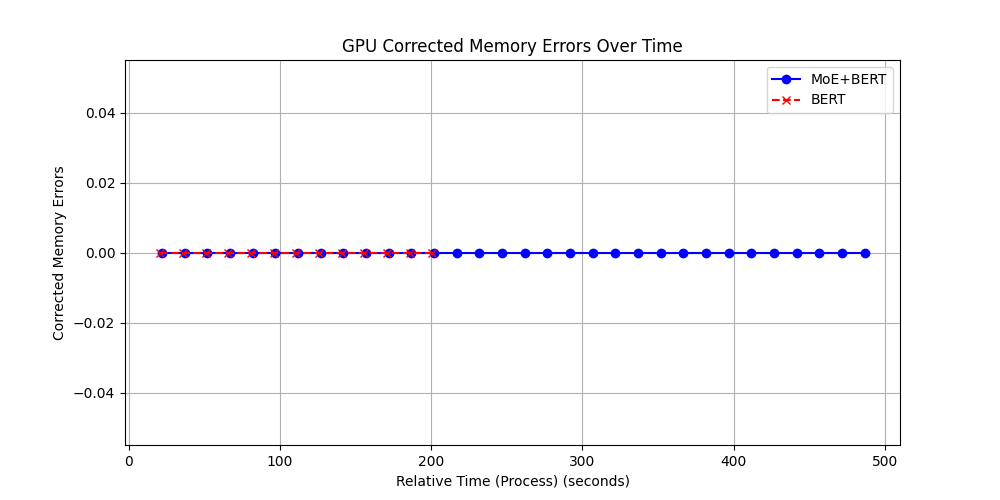}
    \caption{Comparative GPU performance metrics between \textbf{MoE+BERT} (green) and \textbf{BERT} (blue) architectures. 
        Key findings show: (1)~39\% lower power consumption (116W vs~191W), etc.}
    \label{fig12}
\end{figure*}
\endgroup        
\paragraph{Implementation and Reproducibility Details}

All experiments were conducted in Kaggle notebooks using publicly available GPU resources. The final hybrid expert-enhanced architecture (BERT+MoE models), along with all reported results, were trained on two NVIDIA Tesla T4 GPUs (16\,GB VRAM each) using PyTorch Automatic Mixed Precision. Early-stage experiments and baseline BERT models were partially trained on a single NVIDIA Tesla P100 GPU (16\,GB VRAM).
Hyperparameter optimization was performed using the Optuna framework~\citep{akiba2019optuna} with the Tree-structured Parzen Estimator (TPE) sampler~\citep{Bergstra2011}. The search space included batch sizes of \{8, 16, 32\} per GPU and 3--6 training epochs, while the learning rate was tested in the range of $1\times10^{-5}$ to $3\times10^{-5}$ during hyperparameter search. The optimization process aimed to maximize the weighted F1-score on the validation set. The best configuration identified by Optuna---a batch size of 8 per GPU and 3--4 epochs, depending on the training stage---was used for all final models.

\paragraph{Energy Efficiency and Hardware Performance.}

Hybrid expert-enhanced architecture is efficient with regard to hardware. Dynamic expert routing homes in on just the most important segments of the model for each input. This approach cuts energy consumption by nearly 39\%. 
For tourism platforms in Iran and similar emerging markets that process thousands of reviews daily, this directly translates into significant monthly savings in cloud and electricity costs \citep{thematic_review_ai_energy_2025}. 
This shows that sparsely activated AIs significantly mitigate the growing energy consumption attributed to artificial intelligence. The energy savings are remarkable since the model shows performance reliability. The model can still provide stable clock speeds with active memory, which is contrary to many dense transformer models that offer little efficiency with their speed and reliability. This is the working efficiency we look for in Mixture-of-Experts designs. This working efficiency extends to mobile, hybrid expert-enhanced models since their energy costs are operational, ensuring efficient target natural language processing. This is the first of many steps we expect in energy-sustainable AIs—targeting energy consumption while maintaining model accuracy. Overall, the architecture performs strongly in ABSA for Persian tourism reviews, despite the complexities of the language and data imbalance, aiding tourism institutions to analyze reviews better for the users. This improves tourism services.

\section{Conclusion}\label{sec5}

This study introduced a three-stage ABSA framework for Persian tourism reviews and released the 58,473-review Jabama dataset. The proposed hybrid BERT–MoE model achieved a weighted F1-score of 90.6\%, outperforming baseline architectures. Top-K routing and rectification techniques ensured stable expert utilization and reduced GPU power consumption by 39\%. These results demonstrate the model’s suitability for scalable, energy-efficient ABSA in low-resource languages.

\section{Future Work}\label{sec6}

The new hybrid expert-enhanced model performs well on aspect-based sentiment analysis (ABSA) for Persian tourism reviews; however, there are still opportunities for advancement. Detecting sub-aspects of reviews (e.g., `kitchen facilities’ or security systems under `amenities’) using BiO tagging is one way to improve the accuracy of sentiment analysis, ultimately assisting tourism services on Jabama and similar platforms. Focus on hyperparameter tuning in the next work to enhance the model and enable better performance in various tourism contexts. The other opportunity we have is the integration of the model with Interpretability frameworks. Classification of attributes using the Kano model (i.e., basic, performance, excitement) or sentiment shifting feature analysis using SHAP may enhance model transparency and usability, improving the overall user experience.
\backmatter

\bibliography{references}

@article{Kwon2025,
  author    = {Kwon, W.},
  title     = {Aspect-based sentiment analysis through zero-shot text classification and impact-asymmetry analysis},
  journal   = {International Journal of Hospitality Management},
  volume    = {133},
  pages     = {104397},
  year      = {2025},
  doi       = {10.1016/j.ijhm.2025.104397}
}

@article{thematic_review_ai_energy_2025,
  author    = {Qudrat-Ullah, Hassan},
  title     = {A Thematic Review of AI and ML in Sustainable Energy Policies for Developing Nations},
  journal   = {Energies},
  volume    = {18},
  number    = {9},
  pages     = {2239},
  year      = {2025},
  doi       = {10.3390/en18092239},
}

@article{Guidotti2025,
  author    = {Guidotti, Dario and Pandolfo, Laura and Pulina, Luca},
  title     = {Discovering sentiment insights: streamlining tourism review analysis with Large Language Models},
  journal   = {Information Technology \& Tourism},
  volume    = {27},
  pages     = {227--261},
  year      = {2025},
  month     = {April},
  day       = {10},
  doi       = {10.1007/s40558-024-00309-9},
  url       = {https://doi.org/10.1007/s40558-024-00309-9}
}

@article{Das2025,
  author    = {Das, P. and Mandal, S. and Nedungadi, P. and Raman, R.},
  title     = {Unveiling sustainable tourism themes with machine learning based topic modeling},
  journal   = {Discover Sustainability},
  volume    = {6},
  number    = {1},
  year      = {2025},
  doi       = {10.1007/s43621-025-01065-4}
}

@article{Sahin2025,
  author    = {Sahin, G. G. and Eyupoglu, C.},
  title     = {Aspect-Based Sentiment Analysis for Hospitality Industry Applications: A Systematic Literature Review},
  journal   = {Advances in Computational Science and Computing},
  volume    = {30},
  number    = {1},
  year      = {2025},
  doi       = {10.2478/acss-2025-0007}
}

@article{Kaveh2025,
  author    = {Kaveh, S. and Safa, R.},
  title     = {Advancing Natural Language Processing for Persian Movie Review Analysis: Roadmap and Opportunities},
  journal   = {Computational Algorithms and Numerical Dimensions},
  volume    = {4},
  number    = {1},
  pages     = {34--47},
  year      = {2025},
  doi       = {10.22105/cand.2024.493386.1168}
}

@article{Khan2025,
  author    = {Khan, L. and Qazi, A. and Chang, H.-T. and Alhajlah, M. and Mahmood, A.},
  title     = {Empowering Urdu sentiment analysis: an attention-based stacked CNN-Bi-LSTM DNN with multilingual BERT},
  journal   = {Complex Intelligent Systems},
  year      = {2025},
  doi       = {10.1007/s40747-024-01631-9}
}

@article{Nooraee2025,
  author    = {Nooraee, M. and Ghaffari, H. and Kermani, F. Z.},
  title     = {Tiny-ParsBERT: an optimized hybrid model for efficient sentiment analysis in Persian texts},
  journal   = {The Journal of Supercomputing},
  year      = {2025},
  doi       = {10.1007/s11227-025-07297-5}
}

@article{AfsheenMaroof2024,
  author    = {Maroof, A. and Wasi, S. and Jami, S. I. and Siddiqui, M. S.},
  title     = {Aspect Based Sentiment Analysis for Service Industry},
  journal   = {IEEE Access},
  volume    = {12},
  pages     = {1--1},
  year      = {2024},
  doi       = {10.1109/ACCESS.2024.3440357}
}

@article{Jiang2024,
  author    = {Jiang, A. Q. and Sablayrolles, A. and Roux, A. and Mensch, A. and Savary, B. and Bamford, C. and Chaplot, D. S. and de las Casas, D. and Bou Hanna, E. and Bressand, F. and Lengyel, G. and Bour, G. and Lample, G. and Lavaud, L. R. and Saulnier, L. and Lachaux, M.-A. and Stock, P. and Subramanian, S. and Yang, S. and Antoniak, S. and Le Scao, T. and Gervet, T. and Lavril, T. and Wang, T. and Lacroix, T. and El Sayed, W.},
  title     = {Mixtral of Experts},
  journal   = {arXiv preprint arXiv:2401.04088},
  year      = {2024},
  doi       = {10.48550/arXiv.2401.04088}
}

@inproceedings{Muradi2024,
  author    = {Muradi, Malika and Hussain, Basit and Rhythm, Ehsanur Rahman and Rasel, Annajiat Alim},
  title     = {A Comparative Study of ParsBERT and mBERT in Emotion Recognition for Dari-Farsi Text with Explainable AI},
  year      = {2025},
  publisher = {Association for Computing Machinery},
  doi       = {10.1145/3723178.3723231}
}

@misc{Xu2024,
  author    = {Xu, C. and Wang, M. and Ren, Y. and Zhu, S.},
  title     = {Enhancing Aspect-based Sentiment Analysis in Tourism Using Large Language Models and Positional Information},
  year      = {2024},
  doi       = {10.48550/arXiv.2409.14997}
}

@inproceedings{Zeng2024,
  author    = {Zeng, Z. and Guo, Q. and Fei, Z. and Yin, Z. and Zhou, Y. and Li, L. and Sun, T. and Yan, H. and Lin, D. and Qiu, X.},
  title     = {Turn Waste into Worth: Rectifying Top-k Router of MoE},
  booktitle = {Proceedings of the 2024 Conference on Empirical Methods in Natural Language Processing},
  address   = {Miami, Florida},
  publisher = {Association for Computational Linguistics},
  pages     = {9363--9375},
  year      = {2024},
  doi       = {10.18653/v1/2024.emnlp-main.739}
}

@misc{Ghafouri2023,
  author    = {Ghafouri, A. and Abbasi, M. A. and Naderi, H.},
  title     = {AriaBERT: A Pre-trained Persian BERT Model for Natural Language Understanding},
  howpublished = {Research Square preprint},
  year      = {2023},
  doi       = {10.21203/rs.3.rs-3558473/v1}
}

@article{Khizar2023,
  author    = {Khizar, H. M. U. and Younas, A. and Kumar, S. and Akbar, A. and Poulova, P.},
  title     = {The progression of sustainable development goals in tourism: A systematic literature review of past achievements and future promises},
  journal   = {Journal of Innovation \& Knowledge},
  volume    = {8},
  number    = {4},
  pages     = {100442},
  year      = {2023},
  doi       = {10.1016/j.jik.2023.100442}
}

@article{Li2023,
  author    = {Li, H. and Yu, B. X. B. and Li, G. and Gao, H.},
  title     = {Restaurant survival prediction using customer-generated content: An aspect-based sentiment analysis of online reviews},
  journal   = {Tourism Management},
  volume    = {96},
  pages     = {104707},
  year      = {2023},
  doi       = {10.1016/j.tourman.2022.104707}
}

@article{Zhao2023,
author = {Guoshuai Zhao and Yiling Luo and Qiang Chen and Xueming Qian},
title = {Aspect-based sentiment analysis via multitask learning for online reviews},
journal = {Knowledge-Based Systems},
volume = {264},
pages = {110326},
issn = {0950-7051},
year = {2023},
doi = {https://doi.org/10.1016/j.knosys.2023.110326},
}

@article{Fedus2022,
  author    = {Fedus, W. and Zoph, B. and Shazeer, N.},
  title     = {Switch Transformers: Scaling to Trillion Parameter Models with Simple and Efficient Sparsity},
  journal   = {arXiv preprint arXiv:2101.03961},
  year      = {2022},
  doi       = {10.48550/arXiv.2101.03961}
}

@article{Mewada2022,
  author    = {Mewada, A. and Dewang, R. K.},
  title     = {SA-ASBA: a hybrid model for aspect-based sentiment analysis using synthetic attention in pre-trained language BERT model with extreme gradient boosting},
  journal   = {The Journal of Supercomputing},
  volume    = {79},
  number    = {5},
  pages     = {5516--5551},
  year      = {2022},
  doi       = {10.1007/s11227-022-04881-x}
}

@misc{Khodaei2022,
  author    = {Khodaei, A. and Bastanfard, A. and Saboohi, H. and Aligholizadeh, H.},
  title     = {Deep Emotion Detection Sentiment Analysis of Persian Literary Text},
  year      = {2022},
  doi       = {10.21203/rs.3.rs-1796157/v1}
}

@article{Park2022,
  author    = {Park, H. and Jeon, H.},
  title     = {The Dynamics of Customer Satisfaction Dimension based on BERT, SHAP, and Kano Model},
  journal   = {IFAC-PapersOnLine},
  volume    = {55},
  number    = {10},
  pages     = {2384--2389},
  year      = {2022},
  doi       = {10.1016/j.ifacol.2022.10.065}
}

@article{Dashtipour2021,
  author    = {Dashtipour, K. and Gogate, M. and Adeel, A. and Larijani, H. and Hussain, A.},
  title     = {Sentiment analysis of Persian movie reviews using deep learning},
  journal   = {Entropy},
  volume    = {23},
  number    = {5},
  pages     = {596},
  year      = {2021},
  doi       = {10.3390/e23050596}
}

@article{AbbasiMoud2021,
  author    = {Abbasi-Moud, Z. and Vahdat-Nejad, H. and Sadri, J.},
  title     = {Tourism recommendation system based on semantic clustering and sentiment analysis},
  journal   = {Expert Systems with Applications},
  volume    = {168},
  pages     = {114324},
  year      = {2021},
  doi       = {10.1016/j.eswa.2020.114324}
}

@article{Farahani2021,
  author    = {Farahani, M. and Gharachorloo, M. and Farahani, M. and Manthouri, M.},
  title     = {ParsBERT: Transformer-based Model for Persian Language Understanding},
  journal   = {Neural Processing Letters},
  volume    = {53},
  number    = {5},
  pages     = {3831--3847},
  year      = {2021},
  doi       = {10.1007/s11063-021-10528-4}
}

@article{Hu2021,
  author    = {Hu, E. J. and Shen, Y. and Wallis, P. and Allen-Zhu, Z. and Li, Y. and Wang, S. and Wang, L. and Chen, W.},
  title     = {LoRA: Low-Rank Adaptation of Large Language Models},
  journal   = {arXiv preprint arXiv:2106.09685},
  year      = {2021},
  doi       = {10.48550/arXiv.2106.09685}
}

@article{Nandwani2021,
  author    = {Nandwani, P. and Verma, R.},
  title     = {A review on sentiment analysis and emotion detection from text},
  journal   = {Social Network Analysis and Mining},
  volume    = {11},
  pages     = {81},
  year      = {2021},
  doi       = {10.1007/s13278-021-00776-6}
}

@article{Rajabi2021,
  author    = {Rajabi, Z. and Valavi, M. R.},
  title     = {A Survey on Sentiment Analysis in Persian: a Comprehensive System Perspective Covering Challenges and Advances in Resources and Methods},
  journal   = {Cognitive Computation},
  volume    = {13},
  pages     = {882--902},
  year      = {2021},
  doi       = {10.1007/s12559-021-09886-x}
}

@article{Paolanti2021,
  author    = {Paolanti, Marina and Mancini, Adriano and Frontoni, Emanuele and Felicetti, Andrea and Marinelli, Luca and Marcheggiani, Ernesto and Pierdicca, Roberto},
  title     = {Tourism destination management using sentiment analysis and geo-location information: a deep learning approach},
  journal   = {Information Technology \& Tourism},
  volume    = {23},
  number    = {2},
  pages     = {241--264},
  year      = {2021},
  doi       = {10.1007/s40558-021-00196-4},
  url       = {https://doi.org/10.1007/s40558-021-00196-4}
}

@article{Jafarian2020,
  title={Exploiting BERT to Improve Aspect-Based Sentiment Analysis Performance on Persian Language},
  author={Hamoon Jafarian and Amirhosein Taghavi and Alireza Javaheri and Reza Rawassizadeh},
  journal={2021 7th International Conference on Web Research (ICWR)},
  year={2020},
  pages={5-8},
  doi    = {10.5121/ijwest.2020.11401}
}

@article{Lepikhin2020,
  author    = {Lepikhin, D. and Lee, H. and Xu, Y. and Chen, D. and Firat, O. and Huang, Y. and Krikun, M. and Shazeer, N. and Chen, Z.},
  title     = {GShard: Scaling Giant Models with Conditional Computation and Automatic Sharding},
  journal   = {arXiv preprint arXiv:2006.16668},
  year      = {2020},
  doi       = {10.48550/arXiv.2006.16668}
}

@article{DeepSentiPers2020,
  author    = {PourMostafa Roshan Sharami, J. and Abbasi Sarabestani, P. and Mirroshandel, S. A.},
  title     = {DeepSentiPers: Novel Deep Learning Models Trained Over Proposed Augmented Persian Sentiment Corpus},
  journal   = {arXiv preprint arXiv:2004.05328},
  year      = {2020},
  doi       = {10.48550/arXiv.2004.05328}
}

@article{Afzaal2019,
  author    = {Afzaal, M. and Usman, M. and Fong, A.},
  title     = {Tourism Mobile App with Aspect-Based Sentiment Classification Framework for Tourist Reviews},
  journal   = {IEEE Transactions on Consumer Electronics},
  pages     = {233--242},
  year      = {2019},
  doi       = {10.1109/TCE.2019.2908944}
}

@misc{Ataei2019,
  author    = {Ataei, T. S. and Darvishi, K. and Minaei-Bidgoli, B. and Eetemadi, S.},
  title     = {Pars-ABSA: An Aspect-based Sentiment Analysis Dataset for Persian},
  year      = {2019},
  doi       = {10.48550/arXiv.1908.01815}
}

@inproceedings{Devlin2019,
  author    = {Devlin, J. and Chang, M.-W. and Lee, K. and Toutanova, K.},
  title     = {BERT: Pre-training of Deep Bidirectional Transformers for Language Understanding},
  booktitle = {Proceedings of the 2019 Conference of the North American Chapter of the Association for Computational Linguistics: Human Language Technologies, Volume 1 (Long and Short Papers)},
  address   = {Minneapolis, Minnesota},
  publisher = {Association for Computational Linguistics},
  pages     = {4171--4186},
  year      = {2019},
  doi       = {10.18653/v1/N19-1423}
}

@inproceedings{akiba2019optuna,
author = {Akiba, Takuya and Sano, Shotaro and Yanase, Toshihiko and Ohta, Takeru and Koyama, Masanori},
title = {Optuna: A Next-generation Hyperparameter Optimization Framework},
year = {2019},
publisher = {Association for Computing Machinery},
doi = {10.1145/3292500.3330701},
booktitle = {Proceedings of the 25th ACM SIGKDD International Conference on Knowledge Discovery \& Data Mining},
pages = {2623–2631},
numpages = {9},
}

@article{MorenoOrtiz2019,
  author    = {Moreno-Ortiz, Antonio and Salles-Bernal, Soluna and Orrequia-Barea, Aroa},
  title     = {Design and validation of annotation schemas for aspect-based sentiment analysis in the tourism sector},
  journal   = {Information Technology \& Tourism},
  volume    = {21},
  number    = {4},
  pages     = {535--557},
  year      = {2019},
  month     = {June},
  day       = {21},
  doi       = {10.1007/s40558-019-00155-0},
  url       = {https://doi.org/10.1007/s40558-019-00155-0}
}

@misc{Shazeer2017,
  author    = {Shazeer, N. and Mirhoseini, A. and Maziarz, K. and Davis, A. and Le, Q. and Hinton, G. and Dean, J.},
  title     = {Outrageously Large Neural Networks: The Sparsely-Gated Mixture-of-Experts Layer},
  year      = {2017},
  note      = {arXiv preprint arXiv:1701.06538},
  doi       = {10.48550/arXiv.1701.06538}
}

@article{Akhtar2017,
author = {Nadeem Akhtar and Nashez Zubair and Abhishek Kumar and Tameem Ahmad},
title = {Aspect based Sentiment Oriented Summarization of Hotel Reviews},
journal = {Procedia Computer Science},
volume = {115},
pages = {563-571},
year = {2017},
doi = {https://doi.org/10.1016/j.procs.2017.09.115}
}

@inproceedings{Liu2015,
author = {Liu, Qian and Gao, Zhiqiang and Liu, Bing and Zhang, Yuanlin},
title = {Automated rule selection for aspect extraction in opinion mining},
year = {2015},
publisher = {AAAI Press},
booktitle = {Proceedings of the 24th International Conference on Artificial Intelligence},
pages = {1291–1297},
numpages = {7},
doi = {10.5555/2832415.2832429}
}

@inproceedings{Sanguinetti2014,
author = {Soujanya Poria and E. Cambria and Lun-Wei Ku and Chen Gui and Alexander Gelbukh},
title = {A Rule-Based Approach to Aspect Extraction from Product Reviews},
year = {2014},
doi = {10.3115/v1/W14-5905}
}

@inproceedings{Bergstra2011,
 author = {Bergstra, James and Bardenet, R\'{e}mi and Bengio, Yoshua and K\'{e}gl, Bal\'{a}zs},
 booktitle = {Advances in Neural Information Processing Systems},
 editor = {J. Shawe-Taylor and R. Zemel and P. Bartlett and F. Pereira and K.Q. Weinberger},
 publisher = {Curran Associates, Inc.},
 title = {Algorithms for Hyper-Parameter Optimization},
 url = {https://proceedings.neurips.cc/paper_files/paper/2011/file/86e8f7ab32cfd12577bc2619bc635690-Paper.pdf},
 volume = {24},
 year = {2011}
}

\clearpage
\begin{appendices}
\section{Detailed Loss Formulations and Routing Mechanisms}
\label{app:loss_routing}
\subsection{Auxiliary Losses}
In MoE models, a gating network chooses which experts handle each input. If we don’t guide it, only a few experts do most of the work while others sit idle. This makes the model less capable and reduces specialization. To counteract this, we incorporate an auxiliary importance loss inspired by load-balancing objectives proposed in GShard \citep{Lepikhin2020} and later adopted in Switch Transformer \citep{Fedus2022}.

\begin{equation}
\mathcal{L}_{\mathrm{aux}} = \lambda_{\mathrm{aux}} \cdot \frac{\mathrm{Var}(u)}{\mathrm{Mean}(u)^2}, \quad \lambda_{\mathrm{aux}} = 0.011822
\end{equation}
where $\text{Var}(0)$ and $\text{Mean}(0)$ compute the variance and mean across experts, and $\lambda_{\text{aux}}$ controls the strength of the regularization. Minimizing this makes the experts share work more equally and become more specialized.

Another approach clearly punishes variations from the perfect uniform distribution, therefore strongly promoting consistent usage of knowledge.
\begin{equation}
u_{\text{uniform}} = \frac{1}{E} \cdot \mathbf{1}_E \qquad
u_{\text{uniform},e} = \frac{1}{E}, \quad e = 1, 2, \ldots, E.
\end{equation}
This is achieved by introducing a mean squared error (MSE) regularization term:
\begin{equation}
\mathcal{L}_{\mathrm{MSE}} = \lambda_{\mathrm{MSE}} \cdot \frac{1}{E} \sum_{e=1}^{E} \left( u_e - \frac{1}{E} \right)^2,
\end{equation}
where $\lambda_{\mathrm{MSE}}$ is the weight of the MSE term.

\subsection{Evaluation of Expert Utilization}

In order to evaluate the equity of expert usage in the MoE model, we calculated the squared Coefficient of Variation (COV$^2$)~\citep{Shazeer2017} over the routing distributions for 10 batches of the test set. 
The baseline softmax routing yielded COV$^2 = 1.5856$, signifying extreme disparity: experts 0, 1, and 4 were entirely sidelined while others monopolized the routing. 
Even with auxiliary losses alone, COV$^2$ reached 2.1406 and 2.0900, indicating catastrophic routing collapse.

\subsection{Intra-GPU Rectification (IR)}

The Intra-GPU Rectification (IR) focuses on dropped tokens that occur when the number of tokens assigned to an expert surpasses the expert’s capacity limit. 
Instead of discarding these tokens or routing them across GPUs (which incurs high communication costs), IR reroutes them to the optimal expert within the same GPU. 
For a token $x_i$ dropped $(k - |R_i|)$ times (where $R_i$ is the set of successfully routed experts from the initial top-$k$ routing), IR assigns it to the highest-scoring local expert $h$ based on routing scores $a_{ih} = w_h^\top x_i$. 
The combined output $o_i$ is then computed as
\begin{equation}
o_i = \frac{\sum_{j \in R_i} e^{a_{ij}} E_j(x_i) + (k - |R_i|) e^{a_{ih}} E_h(x_i)}{\sum_{j \in R_i} e^{a_{ij}} + (k - |R_i|) e^{a_{ih}}},
\end{equation}
where $E_j(x_i)$ and $E_h(x_i)$ are the outputs from the initial top-$k$ experts and the IR expert, respectively. 
The scaling factor $(k - |R_i|)$ enhances the IR contribution in case of multiple drops. 
IR mitigates routing collapse by maintaining local balance, since dropped tokens per GPU are mostly equitable due to data parallelism.

\subsection{Fill-in Rectification (FR)}

Under-utilized experts are originally filled with zero-padding to sustain balanced workloads across GPUs, resulting in redundant computation. 
Fill-in Rectification (FR) replaces this padding with high-scoring tokens that were not selected in the initial top-$k$ routing. 
For each token $x_i$, FR identifies the $(k+1)$-th highest-scoring expert as a candidate. 
Among tokens selecting the same expert, those with the highest routing scores $a_{i(k+1)}$ are prioritized to occupy the padding positions, effectively extending top-$k$ to top-$(k+1)$ while keeping fixed capacity. 
To prevent vanishing gradients for inactivated experts, we adopt the straight-through estimator during back-propagation by treating the normalization denominator as constant:
\begin{equation}
\frac{\partial \mathcal{L}}{\partial a_{ij}} \equiv \frac{\partial \mathcal{L}}{\sum_j g_{ij}} \cdot \frac{\partial (\sum_j g_{ij})}{\sum_j g_{ij}} \cdot \frac{\partial g_{ij}}{\partial a_{ij}},
\end{equation}
where $g_{ij} = e^{a_{ij}} / \sum_m e^{a_{im}}$.

\subsection{Final Effect of IR + FR}

After incorporating both IR and FR (with capacity factor 1.8 and $K{=}3$), COV$^2$ decreased dramatically to 0.0109, indicating an almost uniform distribution of expert utilization (balanced activity $\approx$ [1.0, 1.0, 1.0, 1.0, 1.0, 1.0]).
Figure~\ref{fig7} illustrates the improved specialization and confirms that no single expert dominates while all six sub-networks actively contribute.

\clearpage

\section{Pseudocode for the MoE-based ABSA Algorithm}
\label{app:pseudocode}

\begin{algorithm}
\caption{Pseudocode for Aspect-Based Sentiment Analysis using MoE with BERT}
\label{alg:moe_absa}
\begin{algorithmic}[1]
\Require Pre-trained fine-tuned BERT model path, Test data Excel file

\Statex \textbf{Step 1: Load Model and Data}
\State Load tokenizer and BERT model from the specified path
\State Move model to device (GPU if available) and set to evaluation mode
\State Read test data from Excel file

\Statex \textbf{Step 2: Preprocess Data}
\For{each row in data}
    \State sentence $\gets$ row['review']
    \State aspect $\gets$ row['Category']
    \State label $\gets$ row['sentiment']
    \State Encode sentence and aspect to obtain input\_ids, attention\_mask, and aspect\_embedding
    \State Append to respective lists
\EndFor
\State Stack inputs into tensors

\Statex \textbf{Step 3: Create and Split Dataset}
\State Create TensorDataset from processed tensors
\State Split into training (80\%), validation (10\%), and test (10\%) sets

\Statex \textbf{Step 4: Define Gate Module}
\State Compute logits and probabilities for expert selection using linear layers and softmax

\Statex \textbf{Step 5: Top-K Dispatch}
\State Select top-$k$ experts based on gate scores
\State Dispatch tokens with capacity limits
\State Combine weighted expert outputs

\Statex \textbf{Step 6: Input Rectification (IR)}
\State Reassign dropped tokens to the best local expert
\State Adjust weights and update outputs

\Statex \textbf{Step 7: Fill-In Rectification (FR)}
\State Fill empty expert slots with top-$(k+1)$ candidates
\State Compute and add outputs

\Statex \textbf{Step 8: Define MoE Model}
\State Use BERT to obtain CLS embedding
\State Apply gate for expert selection
\State Perform top-$k$ dispatch, IR, and FR to produce final logits
\end{algorithmic}
\end{algorithm}

\end{appendices}

\end{document}